# CCNETS: A Modular Causal Learning Framework for Pattern Recognition in Imbalanced Datasets

Hanbeot Park[1], Yunjeong Cho[1], Hunhee Kim[1,2,*]

[1]Department of Data Engineering, Pukyong National University, 45 Yongso-ro, Busan 48513, Republic of Korea; hanbuck30@pukyong.ac.kr (H.P.); yjcho0907@pukyong.ac.kr (Y.C.)

[2]Department of Computer Engineering and Artificial Intelligence, Pukyong National University, 45 Yongso-ro, Busan 48513, Republic of Korea

[*]Corresponding Author: h2kim@pknu.ac.kr

## Abstract

Handling class imbalance remains a central challenge in machine learning, particularly in pattern recognition tasks where identifying rare but critical anomalies is of paramount importance. Traditional generative models often decouple data synthesis from classification, leading to a distribution mismatch that limits their practical benefit. To address these shortcomings, we introduce Causal Cooperative Networks (CCNETS), a modular framework that establishes a functional causal link between generation, inference, and reconstruction. CCNETS is composed of three specialized cooperative modules: an Explainer for latent feature abstraction, a Reasoner for probabilistic label prediction, and a Producer for context-aware data synthesis. These components interact through a dynamic causal feedback loop, where classification outcomes directly guide targeted sample synthesis to adaptively reinforce vulnerable decision boundaries. A key innovation, our proposed Zoint mechanism, enables the adaptive fusion of latent and observable features, enhancing semantic richness and decision-making robustness under uncertainty. We evaluated CCNETS on two distinct real-world datasets: Credit Card Fraud Detection dataset, characterized by extreme imbalance (fraud rate < 0.2%), and the AI4I 2020 Predictive Maintenance dataset (failure rate < 4%). Across comprehensive experimental setups, CCNETS consistently outperformed baseline methods, achieving superior F1-scores, and AUPRC. Furthermore, data synthesized by CCNETS demonstrated enhanced generalization and learning stability under limited data conditions. These results establish CCNETS as a scalable, interpretable, and hybrid soft computing framework that effectively aligns synthetic data with classifier objectives, advancing robust imbalanced learning.

## 1. Introduction

Imbalanced datasets represent one of the most persistent and critical challenges in machine learning, with their importance being particularly pronounced in pattern recognition tasks where rare events carry substantial significance. In real-world application domains such as credit card fraud detection, rare disease diagnosis, and fault detection in industrial systems, the minority class often constitutes only a very small fraction of the overall data, resulting in severely skewed distributions. Under such conditions, standard classification models tend to be excessively optimized toward the majority class, leading to situations where overall accuracy remains high while the recall performance for the minority class—where correct predictions are most crucial—is consistently low.

In application domains that require high-stakes decision-making, the cost structure of classification errors is inherently asymmetric. For example, in credit card fraud detection, even a single false negative—namely, failing to identify a fraudulent transaction—can result in substantial financial losses, legal disputes, and reputational damage. In contrast, false positives lead to the unnecessary blocking of legitimate transactions, increased customer dissatisfaction, and higher operational costs. Consequently, the class imbalance problem extends beyond a matter of performance optimization and constitutes a critical challenge directly tied to real-world operational stability.

Recent advances in artificial intelligence (AI), particularly in deep learning-based models, have led to significant achievements in perception, reasoning, and generation tasks. Artificial neural networks (ANNs) have demonstrated strong performance in high-dimensional problems such as image recognition, natural language processing, and timeseries analysis[1]. Generative models, including variational autoencoders (VAEs) and generative adversarial networks (GANs), have attracted considerable attention as powerful tools in data-limited environments by approximating underlying data distributions and synthesizing new samples[2–4].

Among generative models, GANs have given rise to a wide range of variants due to their ability to synthesize high-quality data through an adversarial learning mechanism between a generator and a discriminator[5]. Models such as Deep Convolutional GAN (DCGAN)[6], SinGAN[7], and StyleGAN[8] have improved generation stability and fidelity, demonstrating broad applicability ranging from handwritten character generation[9] and medical image synthesis[10] to pathological data generation[11–14]. Beyond these applications, generative models have been further extended to diverse tasks including audio synthesis[15,16], semi-supervised learning[17,18], text classification[19,20], and cooperative classifier–generator frameworks[21].

However, such generative-based approaches exhibit structural limitations in class imbalanced settings. In most existing studies, generation and classification are treated as independent stages, and the synthesized data often fail to directly reflect the decision boundaries of the classifier. In particular, when minority class samples are extremely scarce, generative models themselves tend to be biased toward the majority class distribution during training, which can further diminish the representational capacity of the minority class and ultimately constrain classification performance, as noted in multiple prior studies[22–28].

To address these limitations, this study proposes Causal Cooperative Networks (CCNETS) [35],

a collaborative framework that integrates generation, reasoning, and reconstruction as interdependent modules rather than treating them as isolated components. CCNETS consists of three modules—Explainer, Reasoner, and Producer—which continuously interact through a closed-loop feedback structure. In this context, the term "causal" does not refer to statistical causal inference or structural causal models (SCMs). Instead, it emphasizes functional causal relationships and recursive influence, wherein the output of one module directly guides the learning dynamics of another. Specifically, through a feedback mechanism in which classification errors act as driving signals for adjusting the data generation process, CCNETS implements an adaptive learning paradigm that leverages errors and uncertainty as informative learning resources.

Through this causal cooperative structure, CCNETS treats generative and discriminative learning not as independent objectives but as a jointly optimized process. The output of the Reasoner is used as a conditioning signal for the Producer, enabling the synthesis of data concentrated on vulnerable regions of the decision space, while the reconstruction loss between the generated samples and the original inputs is fed back to refine the representations learned by both the Explainer and the Producer. This multi-objective loss formulation prevents the modules from being optimized in isolation and encourages the entire system to simultaneously maintain generative consistency and discriminative capability. In addition, the Zoint mechanism, which adaptively integrates latent features with observable features, is designed to regulate performance trade-offs across different feature fusion strategies, thereby supporting stable learning in highly imbalanced environments.

CCNETS draws conceptual inspiration from the modularity and feedback mechanisms of biological cognitive systems, while its implementation is grounded in engineering pragmatism. Prior studies have demonstrated various approaches that employ generative models—particularly GANs—to reconstruct visual stimuli from neural signals[29–32]. These efforts have extended beyond facial image reconstruction to encompass the reconstruction of complex visual stimuli, such as entire movie scenes[33–36]. While these studies illustrate that generative latent spaces can simulate certain aspects of human perception, they are generally data-intensive, rely on unidirectional mappings, and remain constrained to specific tasks. In contrast, CCNETS is designed as a domain-agnostic framework that generalizes cooperative learning principles centered on interaction and feedback between generative and reasoning modules, explicitly targeting realistic environments characterized by limited data availability and the presence of noise.

We evaluate the generality and robustness of the proposed framework using a credit card fraud detection dataset, in which fraudulent transactions account for approximately 0.17% of all samples, as well as the AI4I 2020 Predictive Maintenance Dataset, where failure events constitute approximately 3.39% of the data. The proposed framework is compared against existing classifiers and representative imbalance-handling techniques to assess performance improvements under extreme class imbalance, while the impact of key module configurations on overall learning performance is also analyzed. In addition, interpretability is examined through latent representation visualization, and the quality and practical utility of the generated data, along with the effectiveness of data augmentation using synthesized samples, are comprehensively evaluated.

In summary, CCNETS is a collaborative framework that integrates generative and discriminative learning through a closed-loop feedback structure, redefining pattern recognition in environments characterized by extreme class imbalance and data scarcity. By treating generative models not merely as data augmentation tools but as functional components of the learning process, CCNETS presents an adaptive and interpretable soft computing–based pattern recognition paradigm that extends beyond traditional oversampling approaches.

## 2. Materials and Method

In this study, we introduce Causal Cooperative Nets (CCNETS) [35], a novel framework designed to facilitate causal learning in neural networks through a structured combination of generative and discriminative modeling components. The architecture and source code are publicly available at https://github.com/bxailab/CCNETS (accessed on 25 January 2026), providing full reproducibility. CCNETS comprises three core modules—Explainer, Reasoner, and Producer—each implemented as a customizable neural network that interacts through shared latent representations and feedback-informed learning pathways. The high-level structure of CCNETS is presented in Figure 1.

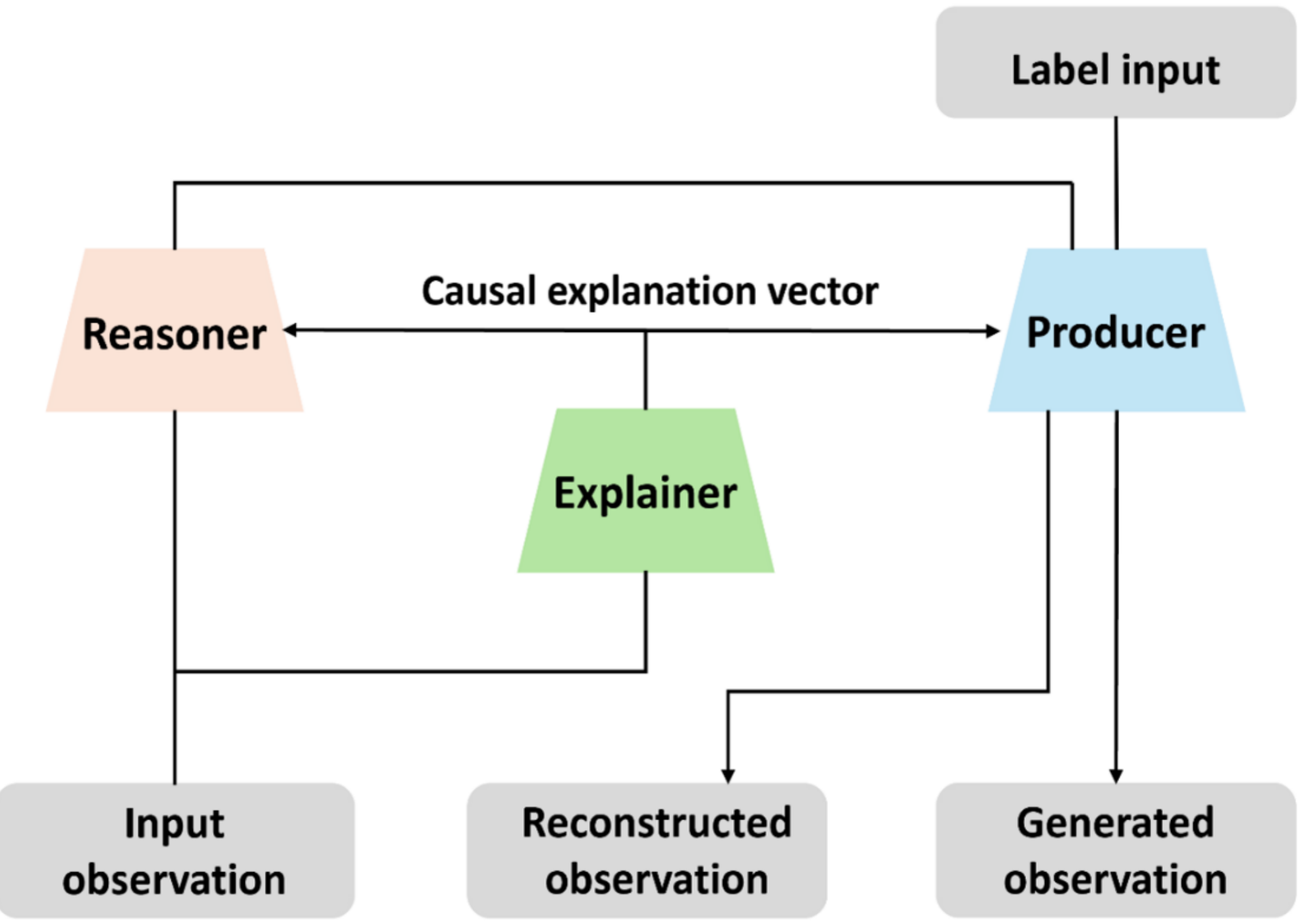

**Figure 1.** Overall structure of CCNETS. CCNETS consists of three interconnected neural network modules: Explainer, Reasoner, and Producer. The Explainer encodes input data into a compact latent space capturing essential features. The Reasoner utilizes the latent space capturing essential features. The Reasoner utilizes the latent representation along with the original input to infer labels through probabilistic reasoning. The Producer generates new or reconstructed input data by integrating the latent space with either provided or inferred labels. These modules collaborate to perform inference, generation, and reconstruction, enabling causal learning across source and target domains.

From a soft computing perspective, CCNETS reflects a hybrid learning architecture that tolerates uncertainty and integrates multiple modules for robust, adaptive decision making CCNETS is designed to perform three key operations: inference, generation, and reconstruction, which are integrated to form a cohesive pipeline for pattern recognition, especially under conditions of data imbalance.

First, in the inference process, the input data are passed into the Explainer, which extracts semantically meaningful features and converts them into a latent space. This latent representation, along with the input data, is then fed into the Reasoner, which infers the most probable label corresponding to the input. Thus, CCNETS enables the prediction of labels from raw observations using both original and abstracted representations.

Second, in the generation process, the input is again passed to the Explainer to generate the latent space and, optionally, the label information. These two outputs—latent encoding and label—are then provided to the Producer, which synthesizes new data that corresponds to the given label. This data generation method allows CCNETS to augment the dataset with class-consistent synthetic samples.

Third, the reconstruction process enables the model to refine its understanding of the input distribution. The input data are processed by the Explainer to yield the latent space, which is used by the Reasoner to predict an inferred label. These two elements—the latent space and inferred label—are then passed to the Producer to generate reconstructed input data. The similarity between the reconstructed and original inputs is minimized through an adaptive loss function during training, enhancing model coherence.

Each module in CCNETS plays a specialized role in the learning cycle. The Explainer transforms raw input data into compact latent embeddings that encapsulate essential feature representations. These embeddings are shared simultaneously with the Reasoner and Producer, ensuring consistent interpretation across modules. The Reasoner uses this latent space, in conjunction with the input data, to perform probabilistic label inference or classification. Meanwhile, the Producer synthesizes new or reconstructed input data by combining the latent space with either true or inferred labels. Through these collaborative operations, CCNETS effectively captures causal relationships between features and labels across source and target domains, using multi-phase loss minimization strategies tailored to each network component.

### 2.1. Explainer

The Explainer module in CCNETS is responsible for encoding input observations into a latent space that captures essential causal features. This encoding process allows subsequent modules—Reasoner and Producer—to perform label inference and data generation effectively. The operation of the Explainer can be formally defined as:

$$\boldsymbol{e} = \boldsymbol{\mathcal{E}}(\boldsymbol{x}) \tag{1}$$

where $x \in R^d$ represents the input observation, $\mathcal{E}(\cdot)$ denotes the Explainer network, and $e \in R^{d'}$ is the resulting latent space embedding.

The internal architecture of the Explainer consists of a series of transformations designed to extract salient features and compress them into a compact latent representation. Specifically, the input passes through the following stages:

(1) Initial transformation:

The input data x are projected into a hidden feature space:

$$\boldsymbol{h_1} = \boldsymbol{\sigma_1}(\boldsymbol{W_1}\boldsymbol{x} + \boldsymbol{b_1}) \tag{2}$$

where $W_1$ and $b_1$ are learnable parameters, and $\sigma_1$ is a non-linear activation function

(typically tanh).

(2) User-customized intermediate network:

The transformed features $h_1$ are processed by a user-defined intermediate network to adapt to specific task requirements:

$$\boldsymbol{h_2} = \boldsymbol{\sigma_2}\left(\text{UserNetwork}(\boldsymbol{h_1})\right) \tag{3}$$

where $\text{UserNetwork}(\cdot)$ is flexibly configurable and $\sigma_2$ commonly uses ReLU activation.

(3) Projection to latent space:

The intermediate features are finally mapped into the latent space:

$$\boldsymbol{e} = \boldsymbol{\sigma_3}(\boldsymbol{W_2}\boldsymbol{h_2} + \boldsymbol{b_2}) \tag{4}$$

where $W_2$ and $b_2$ represent the final transformation parameters, and $\sigma_3$ optionally applies a final activation function.

Through these steps, the Explainer abstracts key information from the raw inputs into a causal explanation vector that is subsequently shared with the Reasoner for label inference and with the Producer for data generation (Figures 2 and 3). Figure 2 illustrates the internal architecture of the Explainer, showing the sequence of transformations from input to latent space, while Figure 3 depicts the data flow from the Explainer to the Reasoner and Producer within CCNETS.

By efficiently encoding the input observations, the Explainer establishes a coherent feature representation that forms the foundation for downstream tasks in causal learning and pattern recognition.

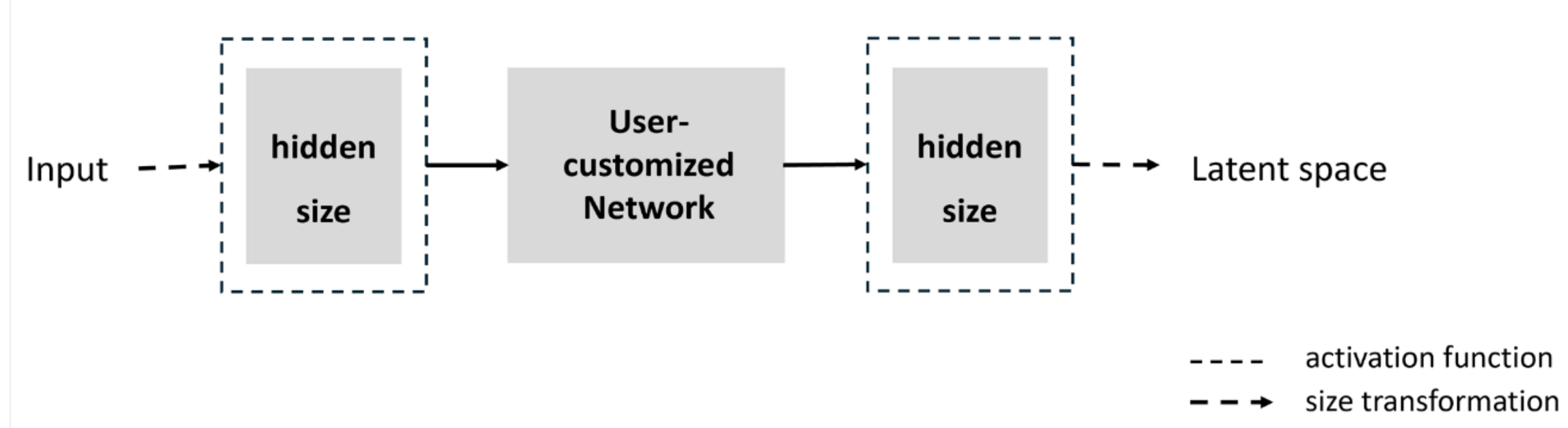

**Figure 2.** Architecture of the Explainer module in CCNETS. This figure illustrates the internal flow of the Explainer network, where input data are first transformed to a hidden size through an activation function. A user-customized network layer is then applied, allowing task-specific feature extraction and flexibility in architecture design. Subsequently, another size transformation and activation function are applied to map the processed features into a final latent space representation. Dashed lines indicate the application of activation functions, while dotted bold lines represent size transformations.

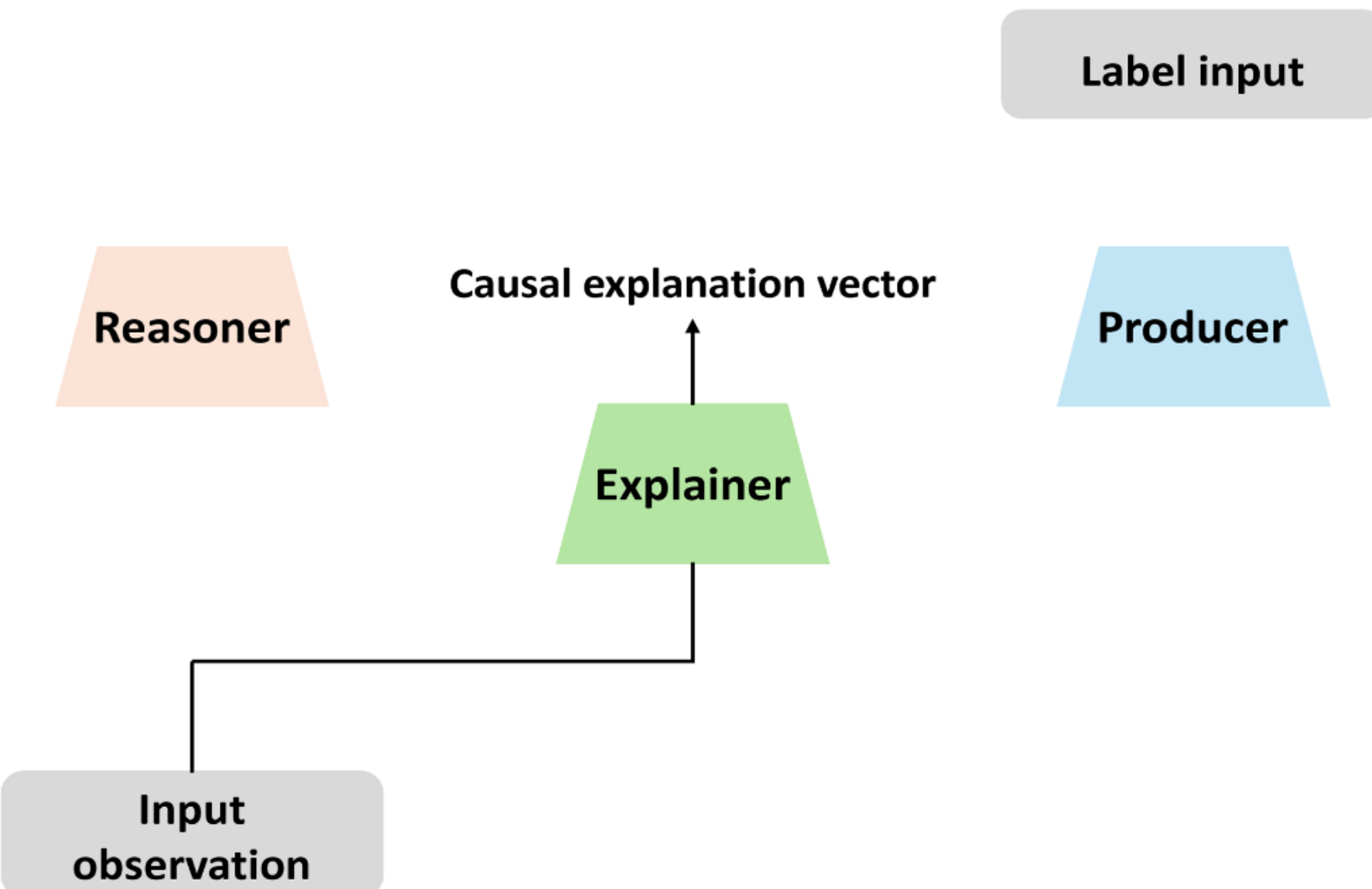

**Figure 3.** Data flow from input observation through the Explainer to Reasoner and Producer modules in CCNETS. This figure shows how the Explainer module processes the input observation to extract a causal explanation vector, which encapsulates the essential features of the input. This vector is then shared with the Reasoner for label inference and the Producer for data generation. The Explainer acts as a critical intermediary that transforms raw input into a structured latent representation used across the subsequent stages of CCNETS.

### 2.2. Zoint Mechanism

The Zoint mechanism is a core operation in CCNETS that integrates the input data and the latent space representations. It provides flexible strategies to combine two information sources before feeding them into the Reasoner and Producer modules, thereby enhancing the network's ability to utilize both raw observations and abstract features.

Formally, let $x \in R^d$ denote the input observation and $e \in R^{d'}$ the latent space vector generated by the Explainer. Since the original input and the latent vector may differ in dimensionality, they are resized into $x'$ and $e'$ to match a common size before combination. This resizing step ensures compatibility for subsequent feature integration.

The Zoint mechanism operates in three different modes:

(1) None mode (independent usage):

$$\boldsymbol{z} = [\boldsymbol{x}, \boldsymbol{e}] \tag{5}$$

In this mode, the input observation and the latent vector are processed separately without any interaction. They are fed independently into subsequent layers.

(2) Add mode (element-wise averaging):

$$\boldsymbol{z} = \frac{1}{2}(\boldsymbol{x}' + \boldsymbol{e}') \tag{6}$$

In this mode, the resized input and latent vectors are averaged element-wise. This operation is analogous to average pooling and helps to merge the two information sources smoothly.

(3) Cat mode (concatenation):

$$\boldsymbol{z} = \text{Concat}(\boldsymbol{x}', \boldsymbol{e}') \tag{7}$$

In this mode, the resized vectors are concatenated to form a single feature vector. This operation preserves the complete information from both the input and latent representations.

Each mode offers distinct advantages depending on the task complexity and the requirements for functional interaction. In the default configuration of this framework, the add mode is applied to both the Reasoner and the Producer. By maintaining a compact representation, the add mode effectively mitigates the risk of over-parameterization and overfitting, which are common in class imbalance scenarios where data is extremely scarce. Furthermore, experimental analysis revealed that the add mode provides a more stable gradient flow compared to the cat mode, ensuring a smoother and more reliable convergence of the loss function. Consequently, as the add mode demonstrated the optimal balance between performance and training stability in fraud detection tasks with severe data imbalance, it was adopted as the default architecture for the final model.

## 2.3. Reasoner

The Reasoner module in CCNETS is responsible for inferring the label corresponding to the input observation based on the latent space generated by the Explainer and the original input data. It acts as the decision-making component that translates abstracted feature representations into actionable predictions.

The operation of the Reasoner can be formally defined as follows:

First, the input observation $x$ and the latent space vector $e$ are fused through the Zoint mechanism:

$$\boldsymbol{z} = \text{Zoint}(\boldsymbol{x}, \boldsymbol{e}) \tag{8}$$

where $z$ denotes the fused feature vector.

The Reasoner network $\mathcal{R}(\cdot)$ then processes the fused feature vector $z$ to predict the label:

$$\boldsymbol{y}' = \boldsymbol{\mathcal{R}}(\boldsymbol{z}) \tag{9}$$

where $y'$ is the predicted label corresponding to the input.

The internal architecture of the Reasoner involves a sequence of transformations, similar in style to the Explainer but adapted for inference tasks:

(1) Zoint Fusion Layer:

The input and latent vectors are combined according to the selected Zoint mode (None, Add, or Cat).

(2) Initial Transformation:

$$\boldsymbol{h_1} = \boldsymbol{\sigma_1}(\boldsymbol{W_1 z} + \boldsymbol{b_1}) \quad (10)$$

where $W_1$ and $b_1$ are learnable parameters, and $\sigma_1$ is a non-linear activation function such as tanh.

(3) User-customized Feature Extraction:

$$\boldsymbol{h_2} = \boldsymbol{\sigma_2}\big(\text{UserNetwork}(\boldsymbol{h_1})\big) \quad (11)$$

where $\text{UserNetwork}(\cdot)$ is a user-defined intermediate network that can be adapted to specific tasks, and $\sigma_2$ typically employs ReLU activation.

(4) Label Prediction:

$$\boldsymbol{y'} = \boldsymbol{\sigma_3}(\boldsymbol{W_2 h_2} + \boldsymbol{b_2}) \quad (12)$$

where $W_2$ and $b_2$ are the final transformation parameters, and $\sigma_3$ represents an optional final activation function that shapes the prediction output (e.g., sigmoid for binary classification).

Figure 4 shows the detailed layer-by-layer structure of the Reasoner, highlighting how feature fusion and transformation occur sequentially. Figure 5 illustrates the inference process, emphasizing the flow from fused features to predicted labels.

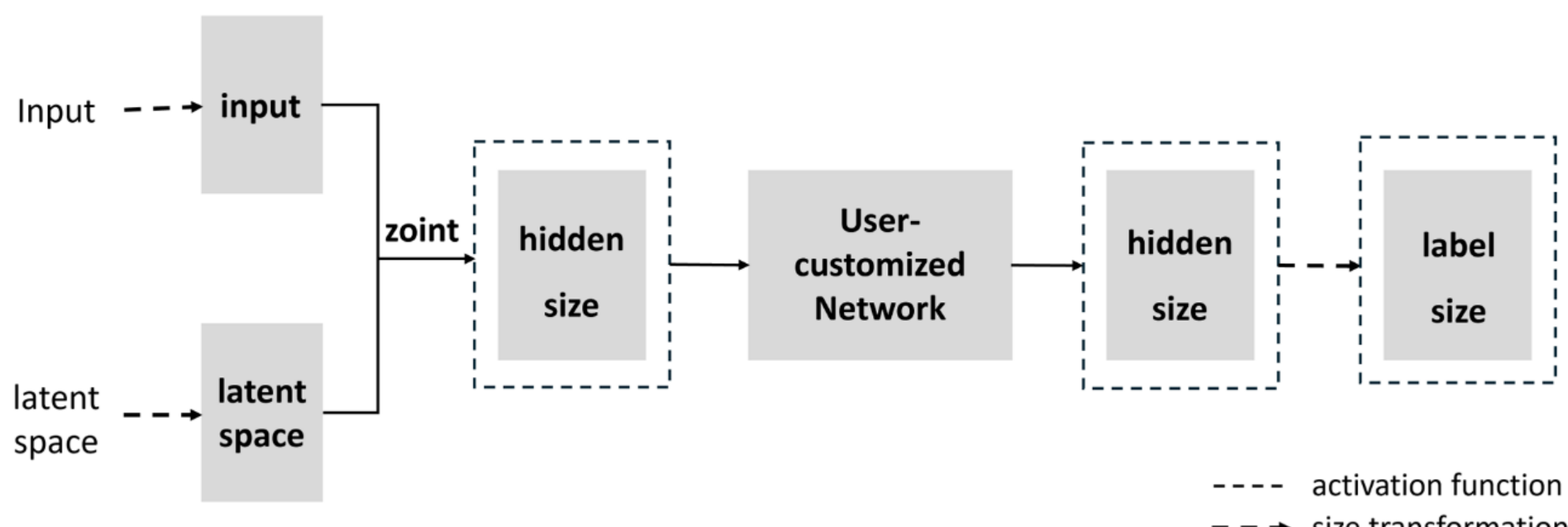

**Figure 4.** Internal architecture of the Reasoner module in CCNETS. This figure illustrates the sequential layers composing the Reasoner, including the Zoint fusion layer, activation layers, user-customized feature extraction, and label prediction layer. The Reasoner combines the latent space vector and input observation to infer labels through a structured transformation process.

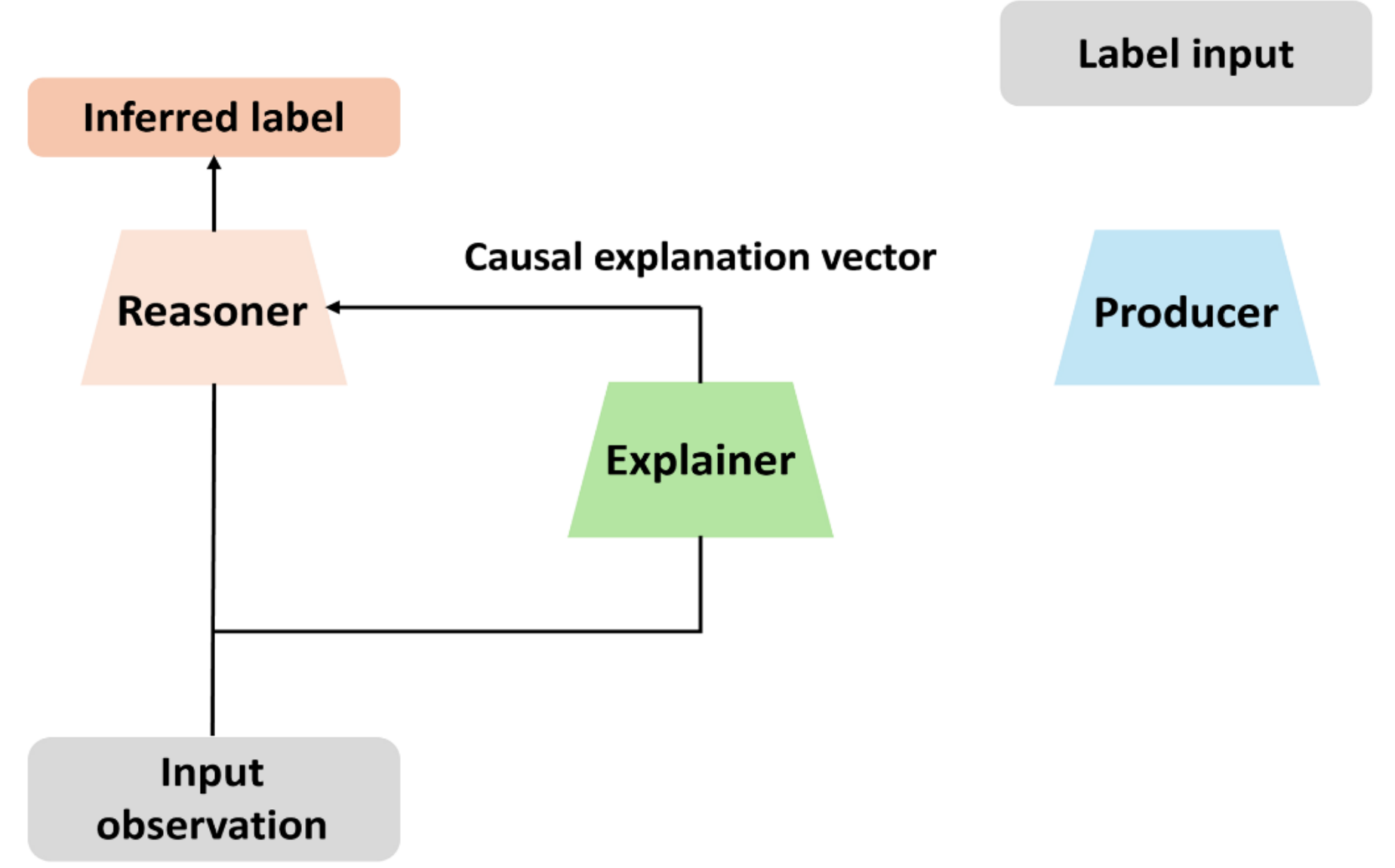

**Figure 5.** Inference process performed by the Reasoner in CCNETS. This figure depicts how the Reasoner receives the latent space and input observation, fuses them using the Zoint mechanism, processes them through intermediate feature extraction, and ultimately infers the predicted label. The Reasoner plays a critical role in translating encoded features into actionable predictions.

Through this structured transformation pipeline, the Reasoner effectively learns to infer accurate labels even in challenging scenarios such as class imbalance, by leveraging both raw observation data and causally extracted latent features.

## 2.4. Producer

The Producer module in CCNETS is responsible for generating new data or reconstructing input observations by utilizing the latent space vector generated by the Explainer and label information, either provided or inferred. The Producer plays a critical role in ensuring that generated data remains consistent with the original data distribution while also enabling augmentation for downstream tasks.

The operation of the Producer can be formally defined as follows:

When using a given label $y$, the latent space vector $e$ and label are fused through the Zoint mechanism:

$$\boldsymbol{z} = \mathrm{Zoint}(\boldsymbol{e}, \boldsymbol{y}) \tag{13}$$

$$\boldsymbol{x}' = \boldsymbol{\mathcal{P}}(\boldsymbol{z}) \tag{14}$$

where $\mathcal{P}(\cdot)$ denotes the Producer network, $z$ is the fused input, and $x'$ is the generated data corresponding to the provided label.

Alternatively, when using the inferred label $y'$ from the Reasoner:

$$\boldsymbol{z}' = \mathrm{Zoint}(\boldsymbol{e}, \boldsymbol{y}') \tag{15}$$

$$\boldsymbol{x}'' = \boldsymbol{\mathcal{P}}(\boldsymbol{z}') \qquad (16)$$

where $x''$ represents the reconstructed data aiming to match the original input observation.

The internal architecture of the Producer mirrors the design principles of the Explainer and Reasoner, comprising sequential transformation stages:

(1) Zoint Fusion Layer:

The latent space vector and label (either provided or inferred) are combined according to the selected Zoint mode.

(2) Initial Transformation:

$$\boldsymbol{h_1} = \boldsymbol{\sigma_1}(\boldsymbol{W_1 z} + \boldsymbol{b_1}) \qquad (17)$$

where $W_1$ and $b_1$ are learnable parameters, and $\sigma_1$ is a non-linear activation function such as tanh.

(3) User-customized Feature Extraction:

$$\boldsymbol{h_2} = \boldsymbol{\sigma_2}\big(\text{UserNetwork}(\boldsymbol{h_1})\big) \qquad (18)$$

where $\text{UserNetwork}(\cdot)$ is a flexible intermediate network, typically employing ReLU activation in $\sigma_2$.

(4) Output Generation:

$$\boldsymbol{x}' = \boldsymbol{\sigma_3}(\boldsymbol{W_2 h_2} + \boldsymbol{b_2}) \qquad (19)$$

where $W_2$ and $b_2$ are the final transformation parameters, and $\sigma_3$ optionally applies a final activation function to control the properties of the generated data.

Figures 6 and 7 illustrate the detailed structure and operational flow of the Producer, emphasizing the generation and reconstruction processes based on causal feature representations.

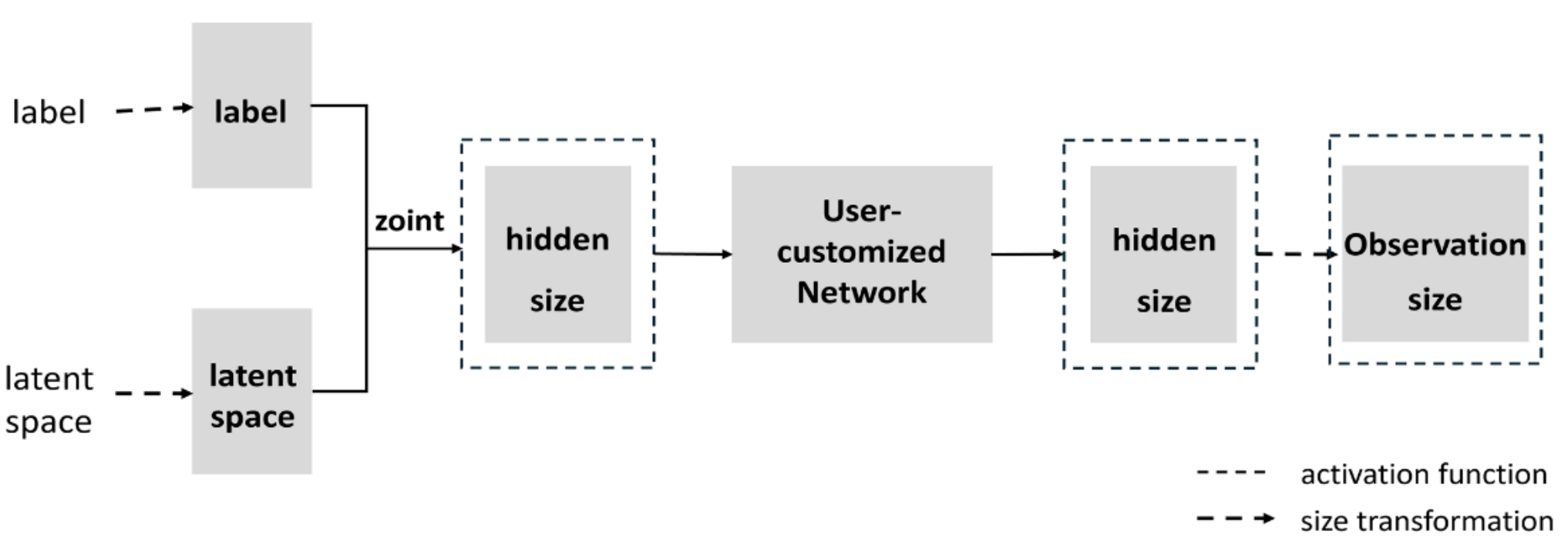

**Figure 6.** Internal architecture of the Producer module in CCNETS. This figure presents the sequential layers of the Producer, illustrating how the latent space and label inputs are fused, processed through feature extraction layers, and ultimately used to reconstruct or generate data corresponding to the specified label.

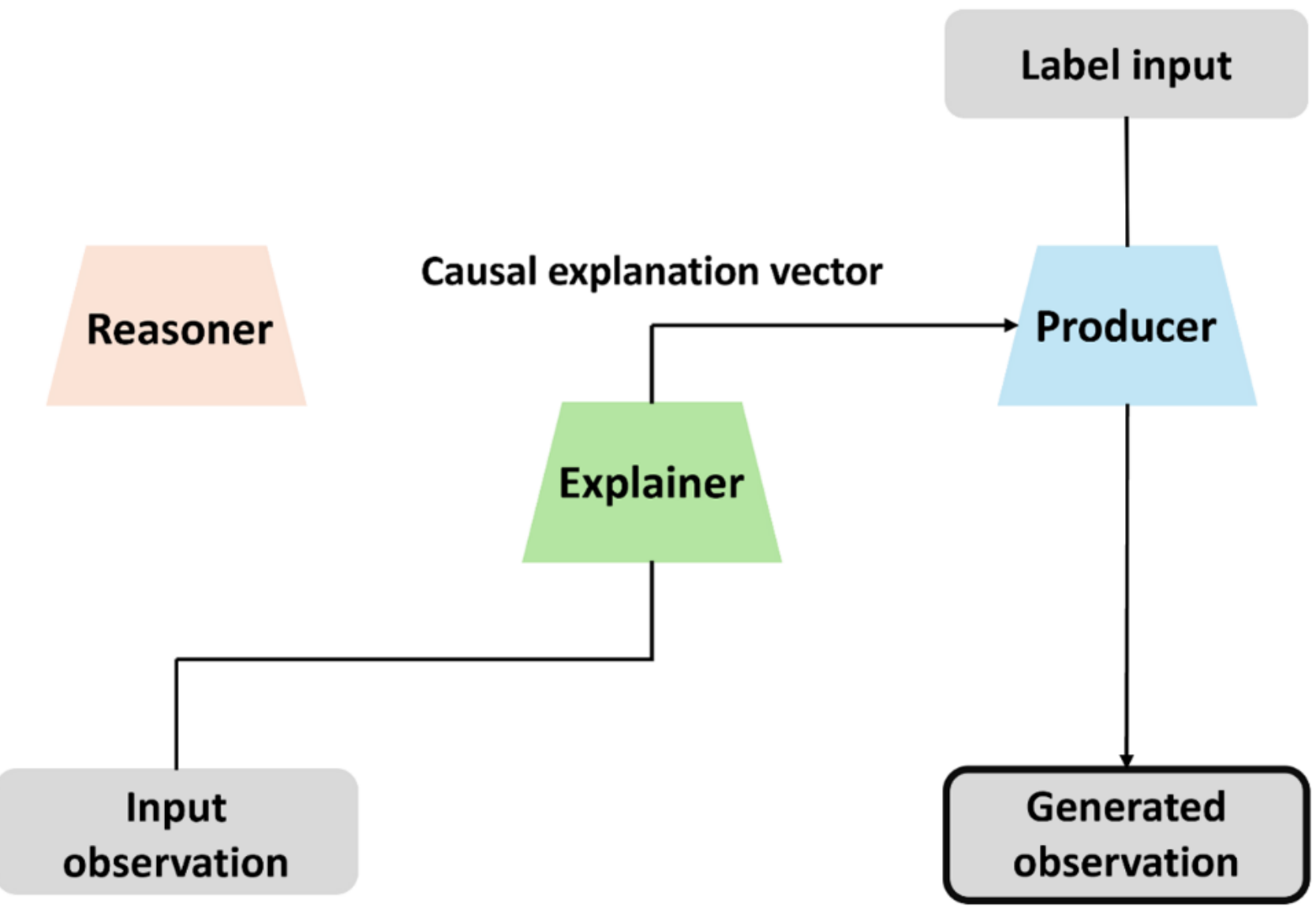

**Figure 7.** Data generation and reconstruction process performed by the Producer in CCNETS. This figure shows how the Producer utilizes the causal latent space vector and label information to generate new data or reconstruct input observations. It highlights the role of the Producer in maintaining coherence between generated data and original data features.

Through this structured approach, the Producer module enables CCNETS not only to enhance data availability through generation but also to reinforce the learning of causal relationships by reconstructing meaningful input patterns.

## 2.5. CCNETS Training and Test

### 2.5.1. Training

The training phase of CCNETS aims to jointly optimize three interconnected modules—Explainer, Reasoner, and Producer—each fulfilling distinct but interrelated roles in causal learning and data generation. The primary objective during training is to ensure that the Explainer extracts essential causal features, the Reasoner accurately infers labels from these features, and the Producer faithfully reconstructs or generates input data based on the inferred or provided labels.

To achieve this, CCNETS minimizes a carefully designed set of loss functions that capture the intricate relationships among input observations, latent representations, generated data, and inferred labels. These loss functions guide each module not only to perform its individual task but also to maintain consistency and coherence across the entire network.

The training procedure is organized into two main processes: loss reduction and prediction loss calculation. Loss reduction focuses on transforming high-dimensional loss tensors into manageable scalar values suitable for optimization, while prediction loss calculation quantifies specific learning objectives, such as label inference accuracy, generation fidelity, and reconstruction consistency. This two-tiered structure enables CCNETS to balance modular specialization with integrated system-level learning.

**Loss Reduction**

To effectively update high-dimensional loss values, CCNETS employs reduction strategies that condense multi-dimensional loss tensors into scalar or manageable forms. Three reduction modes are defined:

(1) All reduction

To ensure that optimization can be efficiently performed, high-dimensional loss tensors must be reduced to scalar values. All reduction computes the mean across all elements of the loss tensor, providing a global measure of model performance during training.

$$\mathbf{Loss}_{\text{all}} = \frac{\mathbf{1}}{\mathbf{N}} \sum_{\mathbf{i=1}}^{\mathbf{N}} \mathbf{l_i} \tag{20}$$

(2) Layer reduction

Layer reduction averages across the batch dimension while preserving the internal structure of the features. It is particularly useful for analyzing layer-wise behavior during training.

$$\mathbf{Loss}_{\text{layer}} = \frac{\mathbf{1}}{\mathbf{B}} \sum_{\mathbf{i=1}}^{\mathbf{B}} \mathbf{l_i} \tag{21}$$

(3) Batch reduction

Batch reduction flattens non-batch dimensions and averages across each batch entry, helping to stabilize learning by focusing on batch-level performance.

$$\mathbf{Loss}_{\text{batch}} = \frac{\mathbf{1}}{\mathbf{B}} \sum_{\mathbf{i=1}}^{\mathbf{B}} \left( \frac{\mathbf{1}}{\mathbf{D}} \sum_{\mathbf{j=1}}^{\mathbf{D}} \mathbf{l_{ij}} \right) \tag{22}$$

where N is the total number of elements, B is the batch size, and D is the flattened dimension.

**Prediction Loss**

During training, CCNETS evaluates three types of prediction losses:

(1) Inference loss

Inference loss evaluates how accurately the system can reconstruct inputs based on inferred labels, reflecting the quality of the Reasoner's label predictions.

$$\boldsymbol{\mathcal{L}}_{\textbf{inference}} = \text{Loss}(\textbf{sg}[\mathbf{X}'], \mathbf{X}'') \quad (23)$$

The operator sg[·] represents the stop-gradient operation, which prevents backpropagation through the target term to ensure stable cooperative learning.

(2) Generation loss

Generation loss measures the fidelity of generated data relative to the original input, indicating how well the Producer can synthesize data from original labels.

$$\boldsymbol{\mathcal{L}}_{\text{generation}} = \text{Loss}(\mathbf{X}, \mathbf{X}') \quad (24)$$

(3) Reconstruction loss

Reconstruction loss assesses the system's ability to recover the original input from the inferred labels, ensuring consistency between generation and inference processes.

$$\boldsymbol{\mathcal{L}}_{\text{reconstruction}} = \text{Loss}(\mathbf{X}, \mathbf{X}'') \quad (25)$$

Each loss type targets different aspects of data quality and coherence within the CCNETS architecture.

**Model Loss**

To ensure stable convergence and distinct role allocation within the CCNETS framework, we employ an alternating optimization strategy where each module utilizes a specific cooperative loss function. During the training of a specific component, the gradients of the other modules are detached, allowing for focused parameter updates.

Based on $\mathcal{L}_{\text{inference}}$, $\mathcal{L}_{\text{generation}}$, $\mathcal{L}_{\text{reconstruction}}$ terms, the objective functions are formulated to anchor specific tasks against stable baselines:

(1) Explainer loss

The Explainer is trained to optimize the latent space e such that the combination of inference consistency and generation quality aligns with the reconstruction baseline. By minimizing the discrepancy between $(\mathcal{L}_{\text{inference}} + \mathcal{L}_{\text{generation}})$ and the detached $\mathcal{L}_{\text{reconstruction}}$, the Explainer extracts robust features that support both downstream tasks without diverging from the manifold of the original data.

$$\boldsymbol{\mathcal{L}}_{\textbf{explainer}} = \boldsymbol{L}_{\boldsymbol{error}}\left(\left(\boldsymbol{\mathcal{L}}_{\textbf{inference}} + \boldsymbol{\mathcal{L}}_{\text{generation}}\right), \textbf{sg}[\boldsymbol{\mathcal{L}}_{\text{reconstruction}}]\right) \quad (26)$$

where $L_{error}$ denotes the chosen distance metric (e.g., MSE or L1 loss) utilized to minimize the discrepancy between the active loss terms and the target constraints.

(2) Reasoner loss

The Reasoner aims to predict a label $\hat{y}$ that yields a reconstruction faithful to both the original input X and the ideal generation $X'$. The objective is to minimize the sum of reconstruction and inference losses $(\mathcal{L}_{\text{reconstruction}} + \mathcal{L}_{\texttt{inference}})$ targeting the detached ground-truth generation error $\text{sg}[\mathcal{L}_{\texttt{generation}}])$ as a regularization anchor. This forces the Reasoner to infer labels that are functionally equivalent to the ground truth in the context of data synthesis.

$$\mathcal{L}_{\textbf{reasoner}} = \boldsymbol{L}_{\boldsymbol{error}}((\mathcal{L}_{\text{reconstruction}} + \mathcal{L}_{\textbf{inference}}), \textbf{sg}[\mathcal{L}_{\textbf{generation}}]) \tag{27}$$

(3) Producer loss

The Producer focuses on high fidelity data synthesis from both ground truth and inferred contexts. Its objective is to minimize the aggregate generation and reconstruction errors $(\mathcal{L}_{\text{reconstruction}} + \mathcal{L}_{\text{generation}})$, utilizing the detached inference consistency gap $\text{sg}[\mathcal{L}_{\texttt{inference}}]$ as a regularization anchor. This ensures that the generator learns to produce realistic samples that faithfully reflect the input distribution, regardless of whether the conditioning label comes from the dataset or the Reasoner.

$$\mathcal{L}_{\textbf{producer}} = \boldsymbol{L}_{\boldsymbol{error}}((\mathcal{L}_{\text{reconstruction}} + \mathcal{L}_{\text{generation}}), \textbf{sg}[\mathcal{L}_{\textbf{inference}}]) \tag{28}$$

Each network component independently optimizes its parameters, ensuring that the entire system converges toward a stable equilibrium state.

Figure 8 visualizes the evolution of these prediction losses during training, highlighting the gradual improvement of the model as each loss component converges.

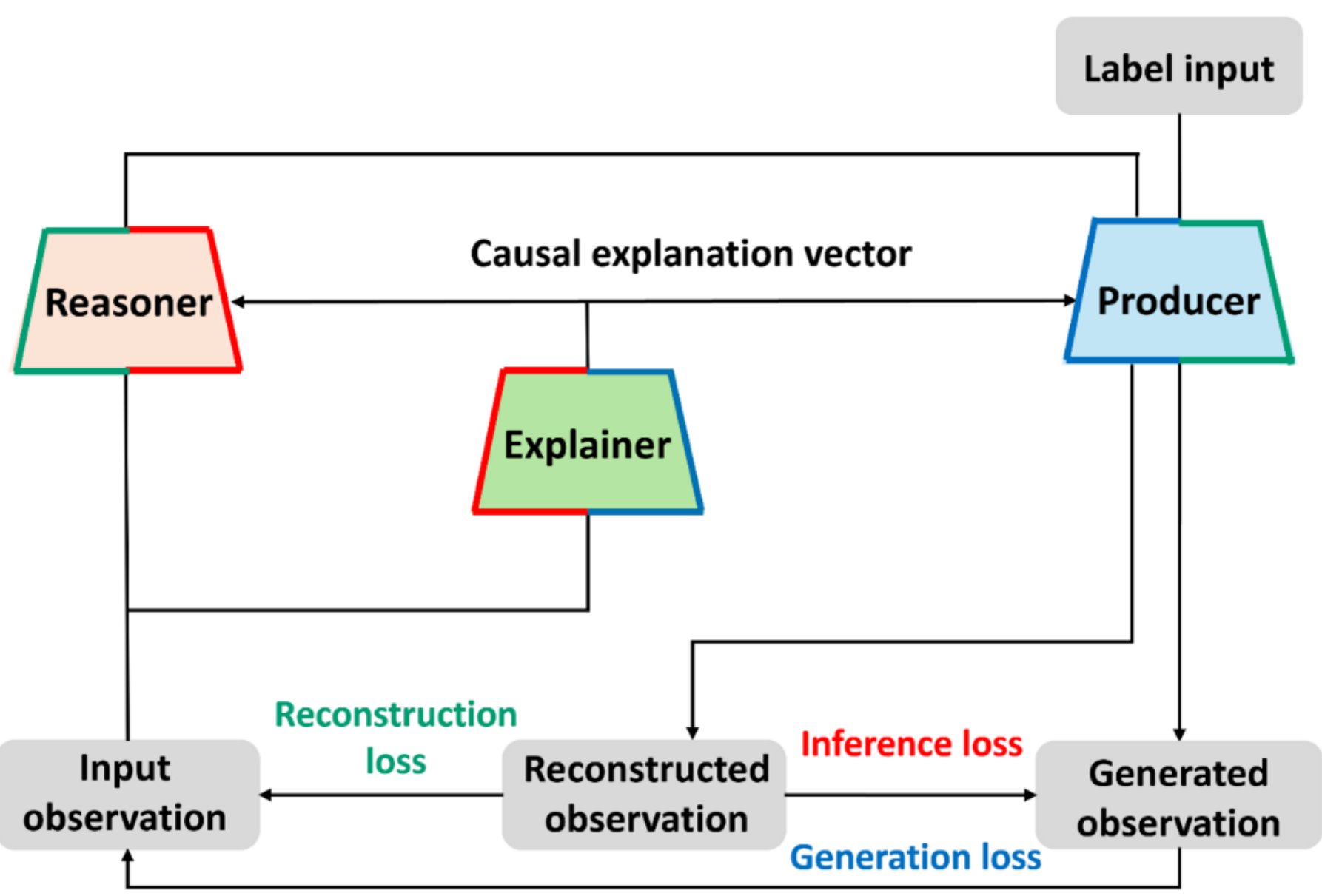

**Figure 8.** Prediction loss components of CCNETS during training [35]. This figure visualizes the three types of prediction losses—inference loss, generation loss, and reconstruction loss—within CCNETS during the training phase. Each loss type measures different aspects of model performance: inference loss evaluates the consistency between generated and reconstructed data, generation loss measures the fidelity of generated data to the original input, and reconstruction loss assesses how well the system recovers the original input from inferred labels.

### 2.5.2. Testing

The testing phase of CCNETS evaluates the generalization ability of the trained networks—Explainer, Reasoner, and Producer—on unseen data. Unlike the training phase, which focuses on optimizing the networks through loss minimization, the test phase measures how well the model can infer labels and maintain data reconstruction coherence without parameter updates. During testing, the input test data $x_{\text{test}}$ are first processed by the trained Explainer network to generate the latent space:

$$\boldsymbol{e}_{\text{test}} = \boldsymbol{\mathcal{E}}(\boldsymbol{x}_{\text{test}}) \tag{29}$$

where $\mathcal{E}(\cdot)$ denotes the Explainer trained during the training phase.
Subsequently, the latent space $e_{\text{test}}$ and the input $x_{\text{test}}$ are fused via the Zoint mechanism, and the Reasoner predicts the inferred label:

$$\boldsymbol{y}'_{\text{test}} = \boldsymbol{\mathcal{R}}\big(\text{Zoint}(\boldsymbol{x}_{\text{test}}, \boldsymbol{e}_{\text{test}})\big) \tag{30}$$

where $\mathcal{R}(\cdot)$ represents the trained Reasoner.

The Producer is not directly involved in the label inference process during testing; however, its previously learned ability to reconstruct input observations from inferred labels contributes to evaluating system coherence if needed.

The main evaluation criteria in the testing phase include:

• Inference performance: The accuracy and reliability of label predictions made by the Reasoner.

• Consistency of latent representations: The coherence between the features extracted by the Explainer and the predicted labels

• Loss behavior: The behavior of inference, generation, and reconstruction losses without training updates, indicating how well the model generalizes to new data.

Specifically, the model's testing loss values are computed by feeding the test data through the fixed Explainer and Reasoner, then calculating the same prediction losses defined during training (inference loss, generation loss, reconstruction loss). No gradient updates or parameter adjustments are performed during this phase.

### 2.5.3. Dataset Description

(1) Credit Card Fraud Detection

This study uses the "Credit Card Fraud Detection" dataset published by the ULB Machine Learning Group, hosted on Kaggle under the Database Contests License (DbCL) v1.0 (https://www.kaggle.com/datasets/mlg-ulb/creditcardfraud (accessed on 15 February 2024)). The dataset used for evaluating CCNETS is a real-world credit card fraud detection dataset, which presents a highly challenging imbalanced classification task. This dataset captures credit card transactions made by European cardholders over a two-day period, consisting of a total of

284,807 transactions.

Each transaction record contains 30 numerical variables, most of which have been transformed via Principal Component Analysis (PCA) for confidentiality and dimensionality reduction purposes. Specifically, 28 variables are PCA-transformed anonymized features, while the remaining two—'Time' and 'Amount'—are original, non-transformed features.

The dataset is severely imbalanced, with fraudulent transactions representing a very small minority of the total samples. Out of all transactions:

$$\text{Fraud Rate} = \frac{\mathbf{492}}{\mathbf{284{,}807}} \times \mathbf{100} \approx \mathbf{0.172}\% \tag{31}$$

Normal transactions account for approximately 99.828%, and fraudulent transactions account for only 0.172%. Such extreme imbalance makes the task particularly suitable for evaluating the robustness and generalization ability of classification models under real-world conditions. For the experiments, the dataset was randomly shuffled and split into a training set and a test set with a 3:7 ratio. This specific split strategy, utilizing a limited amount of training data (30%), was designed to rigorously evaluate the model's data efficiency. By testing on a large unseen dataset (70%), we demonstrate the model's capability to learn robust decision boundaries even under limited supervision scenarios.

A summary of the dataset characteristics is provided below:

• Total transactions: 284,807;

• Fraudulent transactions: 492 (0.172%);

• Normal transactions: 284,315 (99.828%);

• Feature space: 30 variables (28 PCA-transformed, 2 original);

• Time span: Two days;

• Evaluation Protocol: 5-fold Monte Carlo Cross-Validation with Random Shuffling;

• Data split ratio: 3:7 (train:test), sequential without random shuffling.

(2) Fault Detection

This study uses the "AI4I 2020 Predictive Maintenance Dataset" published by the UCI Machine Learning Repository, available under the Creative Commons Attribution 4.0 International (CC BY 4.0) license (https://archive.ics.uci.edu/dataset/601/ai4i+2020+predictive+maintenance+dataset (accessed on 12 January 2026)). The dataset used for evaluating CCNETS is the AI4I 2020 Predictive Maintenance Dataset, which presents a challenging imbalanced classification task in an industrial context. This dataset captures multivariate sensor readings representing the operating conditions of a milling machine, consisting of a total of 10,000 data points.

Each data point consists of a set of process parameters representing the machine's operating condition. Specifically, the categorical feature 'Type' was transformed via one-hot encoding, while the non-informative identifiers—'UDI' and 'Product ID'—were excluded. The remaining variables, including temperature, speed, and torque, serve as the primary numerical inputs.

The dataset is severely imbalanced, with machine failures representing a very small minority of the total samples. Out of all data points:

$$\text{Fraud Rate} = \frac{\mathbf{339}}{\mathbf{10,000}} \times \mathbf{100} \approx \mathbf{3.39}\%$$

Normal operating states account for approximately 96.61%, and failure events account for only 3.39%. Such extreme imbalance makes the task particularly suitable for evaluating the robustness and generalization ability of fault detection models under realistic industrial conditions.

For the experiments, the dataset was randomly shuffled and split into a training set and a test set with a 3:7 ratio. This specific split strategy, utilizing a limited amount of training data (30%), was designed to rigorously evaluate the model's data efficiency. By testing on a large unseen dataset (70%), we demonstrate the model's capability to learn robust decision boundaries even under limited supervision scenarios.

A summary of the dataset characteristics is provided below:

• Total data points: 10,000;

• Failure events: 339 (3.39%);

• Normal operating states: 9661 (96.61%);

• Feature space: 13 variables (10 numerical sensor readings, 3 one-hot encoded categorical features);

• Time span: One day;

• Evaluation Protocol: 5-fold Monte Carlo Cross-Validation with Random Shuffling;

• Data split ratio: 3:7 (train:test), sequential without random shuffling.

This dataset poses a significant challenge in both detecting rare events and maintaining high precision and recall, making it an ideal benchmark for validating the effectiveness of CCNETS in handling highly imbalanced and anonymized data environments.

**2.6. Parameters**

The performance of CCNETS is influenced by the careful selection of hyperparameters across its Explainer, Reasoner, and Producer modules. This section describes the optimization, loss configuration, network structure, and fusion mechanisms adopted for training and testing.

(1) Optimization Settings

Each module was trained using the AdamW optimizer with a learning rate $\eta$ of $2 \times 10^{-4}$ and a weight decay λ of 0.01. The optimizer updates parameters θ according to:

$$\boldsymbol{\theta}_{t+1} = \boldsymbol{\theta}_t - \boldsymbol{\eta} * \left(\frac{\widehat{m}_t}{\sqrt{(\widehat{v}_t)} + \epsilon} + \boldsymbol{\lambda}\boldsymbol{\theta}_t\right) \tag{32}$$

where $\hat{m_t}$ and $\hat{v}_t$ are the bias-corrected estimates of the first and second moments, and $\varepsilon$ is a small constant to prevent division by zero, and λ denotes the weight decay coefficient.

A step-based learning rate decay strategy was employed, following:

$$\boldsymbol{\eta_t = \eta_0 * \gamma^{\lfloor t/stepsize \rfloor}} \quad (33)$$

with a decay rate $\gamma = 0.99954$ and a step size $t$ of 10 epochs.

(2) Loss Settings

The L1 loss function was adopted for both prediction and model losses due to its robustness against outliers and promotion of sparsity. Loss reduction was primarily performed by averaging across all elements, ensuring stable optimization.

(3) Structural Parameters

Distinct inner network structures were assigned to each module:

• Explainer: Multilayer Perceptron (MLP);

• Reasoner: DeepFM (Deep Factorization Machine);

• Producer: ResMLP (Residual MLP).

Each module consisted of three layers with a hidden size of 256 units. The latent space was set to 128 dimensions, and the label size was set to 1 for binary classification tasks.

(4) Fusion Mechanism

For both the Reasoner and Producer, the add mode was employed in the Zoint mechanism, enabling a balanced integration of latent and observable features without feature dimension explosion.

A summary of all hyperparameter settings is provided in Table 1. These parameters were empirically selected to balance model performance, convergence stability, and generalization ability across highly imbalanced datasets.

**Table 1.** Hyperparameters of CCNETS components. This table summarizes the hyperparameters individually optimized for the Explainer, Reasoner, and Producer networks of CCNETS, including learning rate, optimizer configuration, and network architecture. These settings are selected to optimize performance on the fraud detection task.

| | **Explainer** | **Reasoner** | **Producer** |
|---|---|---|---|
| Learning rate | $2 \times 10^{-4}$ | $2 \times 10^{-4}$ | $2 \times 10^{-4}$ |
| gamma | 0.99954 | 0.99954 | 0.99954 |
| Beta1 | 0.9 | 0.9 | 0.9 |
| Beta2 | 0.999 | 0.999 | 0.999 |
| Step size | 10 | 10 | 10 |
| optimizer | AdamW | AdamW | AdamW |
| Prediction loss type | L1 | L1 | L1 |
| Model loss type | L1 | L1 | L1 |
| Prediction loss reduction | All | All | All |
| Model loss reduction | None | None | None |

| Inner network | MLP | deepfm | resMLP |
|---|---|---|---|
| Reasoner joint type | | add | |
| Producer joint type | | | add |
| Inner network number of layer | 3 | 3 | 3 |
| Observe size | Input feature | Input feature | Input feature |
| Label size | 1 | 1 | 1 |
| Explain size | 128 | | |
| Final layer’s activation function | none | none | none |
| Hidden size | 256 | 256 | 256 |

The full CCNETS model includes approximately 1M trainable parameters across all modules and converges within 100 epochs on the given dataset. Training on a single GPU (NVIDIA RTX 4090, Santa Clara, CA, USA) typically completed in under 7 min, indicating its feasibility for real-world deployment scenarios.

## 3. Results

### 3.1. Overview of Experiments

In this study, we designed a series of systematic experiments to rigorously evaluate the effectiveness of CCNETS in addressing highly imbalanced classification tasks, including fraud detection and industrial fault detection. The primary objective is to assess the ability of CCNETS to enhance pattern recognition performance compared to traditional approaches and to investigate the impact of its generative mechanisms.

The experimental framework was structured into five sequential stages, each aimed at answering a specific research question:

First, we conducted a comparative analysis between CCNETS and various baseline approaches, including the standalone DeepFM model, autoencoder-based anomaly detectors, and DeepFM integrated with standard imbalance handling techniques. This was to verify whether the novel causal cooperative network structure of CCNETS could provide a significant performance advantage over both conventional classifiers and existing imbalance handling solutions in extremely imbalanced environments.

Second, we conducted an ablation study to investigate the performance variations resulting from different Joint Types utilized in the Reasoner and Producer modules. Specifically, we compared three distinct configurations: add type, cat type, and none type. This analysis was intended to identify the optimal feature integration strategy and to empirically justify the selection of the add type as the default configuration for the CCNETS architecture.

Third, we conducted a qualitative analysis by visualizing the latent space constructed from the features extracted by the Explainer and Reasoner modules. This analysis aimed to verify whether CCNETS effectively captures high-level semantic features and demonstrates interpretability, even within severely imbalanced environments.

Fourth, we examined the effectiveness of data generated by CCNETS in improving supervised learning models. We compared the performance of models trained on original datasets with those trained on datasets generated or reconstructed by CCNETS to validate the quality and utility of the generated data.

Fifth, we investigated the impact of significantly increasing the volume of generated data by tenfold. The goal was to determine whether data augmentation through CCNETS could yield tangible improvements in model generalization and predictive performance, addressing the issue of data scarcity.

Throughout experiments key evaluation metrics included the F1 score, precision, recall. Specifically for Experiments 1 and 2, ROC-AUC and AUPRC were additionally utilized to address imbalanced classification challenges. Furthermore, the loss curves during both training and testing phases were analyzed across the experiments to evaluate the overall learning dynamics and convergence behavior.

To rigorously assess the statistical significance of the performance improvements, we employed the one-sided Wilcoxon signed-rank test. This non-parametric approach was chosen to account for the potential deviation from normality in the distribution of performance metrics across the 5-fold cross-validation. A significance level of $p < 0.05$ was used to determine statistical validity.

This multilayered experimental design facilitates a comprehensive and progressive validation of CCNETS, moving from basic performance benchmarking against existing methods to exploring the benefits of advanced data generation strategies, thereby constructing a cohesive and logical narrative across the results section.

**3.2. Hyperparameter Setting**

To ensure optimal performance of the CCNETS model in the fraud detection task, we carefully tuned the hyperparameters of each component: Explainer, Reasoner, and Producer. Table 1 presents a detailed summary of the hyperparameters configured for each network.

The learning rate was uniformly set to 0.0002 across all three networks, balancing convergence speed and stability. The AdamW optimizer, known for its adaptive moment estimation capabilities, was employed with parameters Beta1 = 0.9, Beta2 = 0.999, and a weight decay of 0.01. Additionally, a decay rate gamma of 0.99954 and a step size of 10 epochs were set for the learning rate schedule to facilitate efficient learning.

For the prediction loss, the L1 loss function was used consistently across the Explainer, Reasoner, and Producer networks. This choice aimed to robustly handle outliers, which are prevalent in highly imbalanced datasets. Model loss types were also based on the L1 criterion.

Regarding loss reduction methods, the prediction losses were aggregated using the “All” reduction strategy, meaning an average was taken across all elements, while no additional reduction was applied to the model losses to preserve their detailed structure during optimization.

Each network utilized a specific internal structure optimized for its role: the Explainer employed an MLP architecture, the Reasoner incorporated a deep factorization machine (DeepFM), and the Producer used a residual MLP (ResMLP) design. These architectures were selected to maximize feature extraction, inference, and data generation capabilities, respectively.

The internal structure of each network consisted of three layers, ensuring sufficient capacity without overfitting. Observed feature size, label size, and explained latent size were all standardized at Input feature size, 1, and 128 dimensions, respectively. The final activation functions were chosen based on the module’s output requirements: the Explainer applied a sigmoid activation, whereas the Reasoner and Producer omitted final activations to retain raw output values.

The hidden layer size was uniformly set to 256 across all components. This dimension was empirically selected to mitigate the risk of overfitting, a common pitfall when applying over-parameterized networks (e.g., 512 or 1024 dimensions) to highly imbalanced data. While baseline models share this configuration to ensure a fair comparison, CCNETS maximizes parameter efficiency through its cooperative modular design.

These carefully selected hyperparameters, as summarized in Table 1, were critical in achieving stable training, effective convergence, and superior performance in subsequent experiments, ensuring the reliable evaluation of CCNETS against conventional approaches.

### 3.3. Experiment 1: Comparison with Traditional Generative Models

The first experiment benchmarks the classification performance of CCNETS against baseline models, including the standalone DeepFM classifier, traditional autoencoder-based anomaly detectors, and DeepFM integrated with various imbalance handling techniques.

Table 2 summarizes the results obtained from the Credit Card Fraud Detection dataset. CCNETS achieved an F1 score of 0.829, significantly outperforming the autoencoder-based method (0.800) and other DeepFM-based strategies ($p < 0.05$). Notably, CCNETS also recorded an AUPRC of 0.808, surpassing all baseline models by a statistically significant margin ($p < 0.05$). This superior performance in both F1 and AUPRC suggests that CCNETS maintains an optimal balance between precision and recall, a critical factor for effective financial fraud detection.

**Table 2.** Performance comparison of CCNETS against baseline models and various imbalance handling strategies on the Credit Card Fraud Detection dataset. The evaluation metrics include Precision, Recall, F1-score, ROC-AUC, and AUPRC. The comparison includes the standalone DeepFM model, Autoencoder-based approaches, and DeepFM integrated with standard imbalance handling techniques.

| | F1 | Precision | Recall | AUPRC | ROC-AUC |
|---|---|---|---|---|---|
| CCNETS | **0.829** | **0.847** | 0.812 | **0.808** | 0.928 |
| Auto Encoder | 0.800 | 0.834 | 0.772 | 0.775 | **0.953** |
| DeepFM | 0.796 | 0.818 | 0.776 | 0.769 | 0.945 |
| DeepFM(CTGAN) | 0.760 | 0.798 | 0.739 | 0.730 | 0.932 |
| DeepFM(SMOTE) | 0.784 | 0.783 | 0.787 | 0.750 | 0.946 |
| DeepFM(Focal) | 0.784 | 0.814 | 0.762 | 0.771 | 0.943 |
| DeepFM(Weighted) | 0.411 | 0.287 | **0.844** | 0.572 | 0.935 |

Note: Boldface marks the best performance in each column.

Furthermore, Table 3 presents the performance comparison on the AI4I 2020 Predictive Maintenance dataset. In this industrial fault detection task, CCNETS achieved an F1 score of 0.984, demonstrating top-tier performance among all baselines. While this F1 score is comparable to that of the autoencoder-based method (0.984), CCNETS exhibited a distinct advantage in other critical metrics. Specifically, it significantly outperformed all baselines in terms of AUPRC (0.977) and ROC-AUC (0.987) ($p < 0.05$). These results confirm that even when standard metrics like F1 reach a saturation point, CCNETS provides more robust and reliable probability estimates for identifying rare machine failure events.

In addition to performance metrics, we analyzed the loss trends during training and testing phases using Figures 9 and 10. Figure 9 illustrates the steady decline in loss values for the Explainer, Reasoner, and Producer networks throughout the training process. Specifically, Explainer loss, Reasoner loss, Producer loss, and prediction losses (inference, generation, reconstruction) all demonstrated consistent downward trends, with high R 2 scores, indicating stable and effective training dynamics.

Figure 10 visualizes the evolution of model and prediction losses during the testing phase. The consistent downward trend across all loss components reflects the robustness of CCNETS. This continuous decrease indicates that the model successfully maintains the relationships learned during training even when encountering new, unseen data, thereby validating its capability to generalize causal patterns beyond the training distribution.

These results collectively validate that CCNETS provides a substantial advantage over traditional generative model approaches, achieving better recall while maintaining a strong balance with precision, alongside stable and reliable learning behavior.

**Table 3.** Performance comparison of CCNETS against baseline models and various imbalance handling strategies on the AI4I 2020 Predictive Maintenance dataset. The evaluation metrics include Precision, Recall, F1-score, ROC-AUC, and AUPRC. The comparison includes the standalone DeepFM model, Autoencoder-based approaches, and DeepFM integrated with standard imbalance handling techniques.

| | F1 | Precision | Recall | AUPRC | ROC-AUC |
|---|---|---|---|---|---|
| CCNETS | **0.984** | 0.997 | **0.973** | **0.977** | **0.987** |
| Auto Encoder | **0.984** | **0.998** | 0.970 | 0.974 | 0.984 |
| DeepFM | 0.977 | 0.996 | 0.959 | 0.970 | 0.983 |
| DeepFM(CTGAN) | 0.980 | 0.990 | 0.972 | 0.974 | 0.985 |
| DeepFM(SMOTE) | 0.962 | 0.972 | 0.954 | 0.970 | 0.983 |
| DeepFM(Focal) | 0.975 | 0.994 | 0.956 | 0.968 | 0.983 |
| DeepFM(Weighted) | 0.960 | 0.963 | 0.958 | 0.968 | 0.984 |

Note: Boldface marks the best performance in each column.

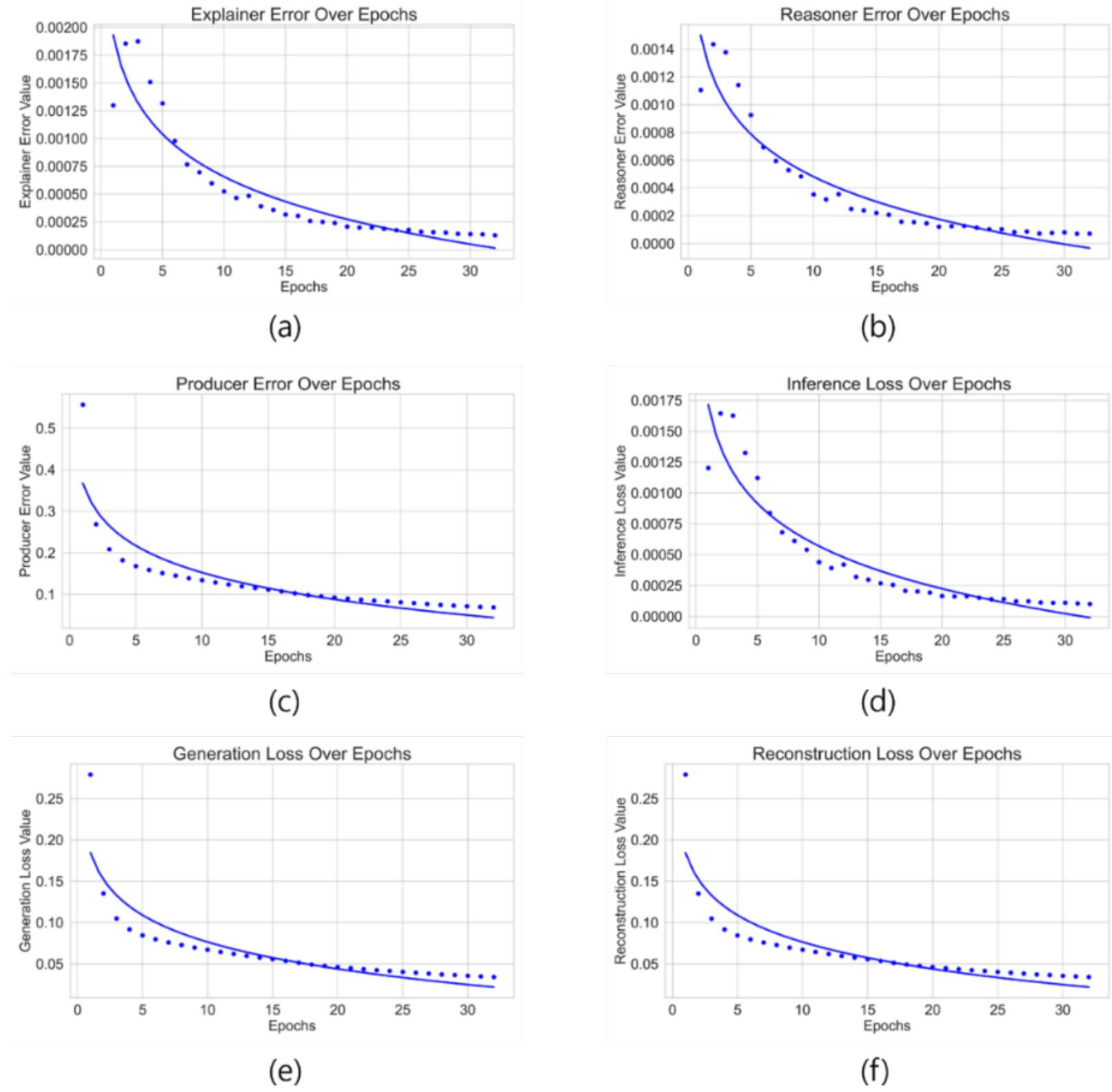

**Figure 9.** Training losses of CCNETS models across 32 epochs. This figure shows the reduction trends of the Explainer loss, Reasoner loss, Producer loss, inference loss, generation loss, and reconstruction loss during CCNETS training using L1 loss. The steady decrease of all losses demonstrates effective learning progress. Specifically, (a) Explainer loss with a score of 0.8534, (b) Reasoner loss with an score of 0.8885, (c) Producer loss with a score of 0.7846, (d) inference loss with a score of 0.8701, (e) generation loss with a score of 0.7877, and (f) reconstruction loss with a score of 0.7869.

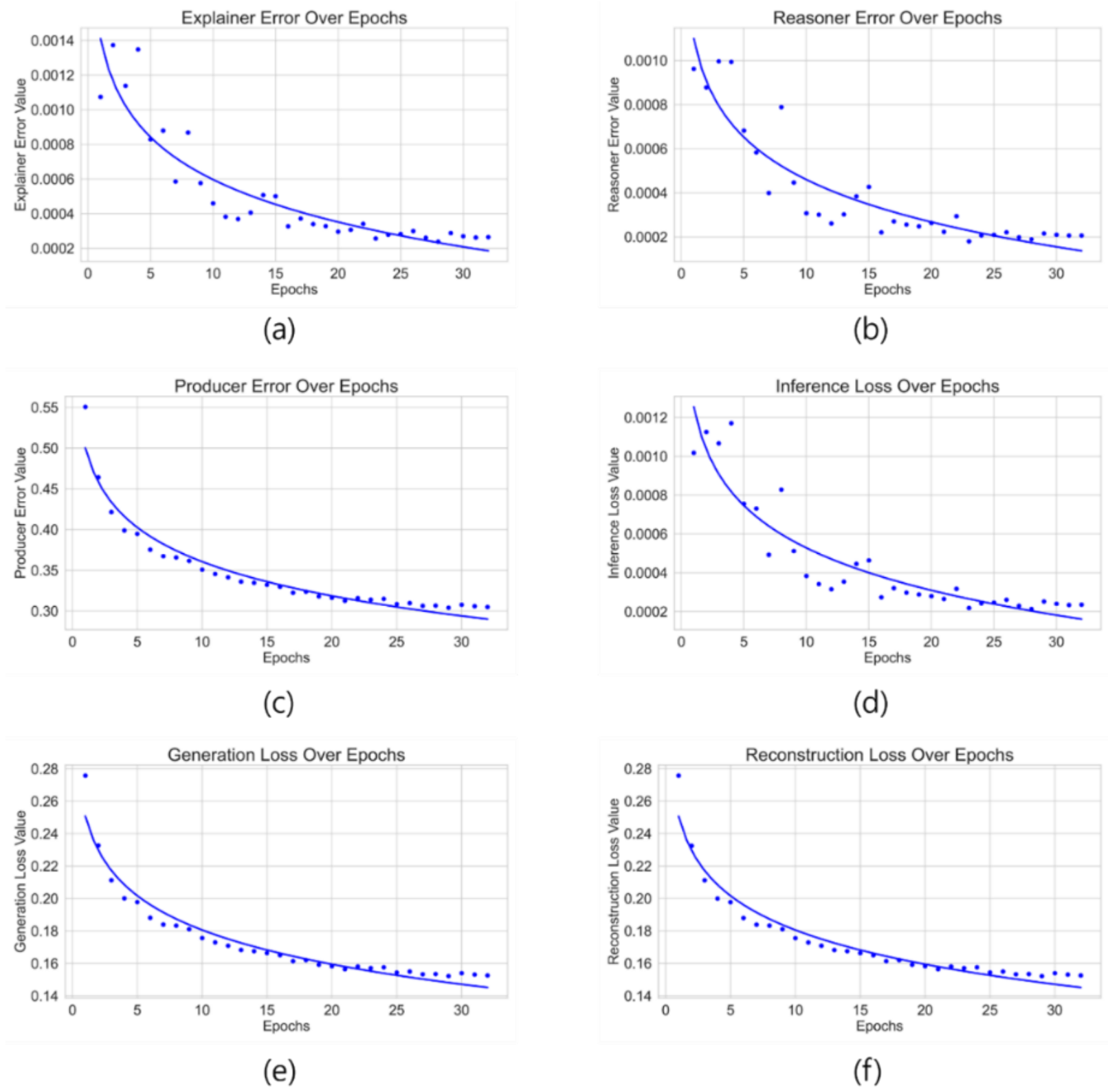

**Figure 10.** Testing losses of CCNETS models across 32 epochs. This figure presents the reduction trends of losses on the test dataset during CCNETS evaluation. The consistent decrease indicates that the model generalizes well from training to unseen data. Specifically, (a) Explainer loss with an $\boldsymbol{R^2}$ score of 0.8359, (b) Reasoner loss with an $\boldsymbol{R^2}$ score of 0.8322, (c) Producer loss with an $\boldsymbol{R^2}$ score of 0.9434, (d) inference loss with an $\boldsymbol{R^2}$ score of 0.8447, (e) generation loss with an $\boldsymbol{R^2}$ score of 0.9443, and (f) reconstruction loss with an $\boldsymbol{R^2}$ score of 0.9440.

### 3.4. Experiment 2: Effectiveness of Generated Data

The second experiment serves as an ablation study to analyze the performance variations of CCNETS configurations with different joint types.

As shown in Table 4, on the Credit Card Fraud Detection dataset, the add type demonstrated superior performance. It achieved statistically significant gains over other types in F1-score, Precision, AUPRC, and ROC-AUC ($p < 0.05$). Although the improvement in Recall was not statistically significant, the ‘add’ type still yielded the highest value among all configurations.

**Table 4.** Performance comparison of CCNETS configurations with different Joint Types on the Credit Card Fraud Detection dataset. The evaluation metrics include Precision, Recall, F1-score, ROC-AUC, and AUPRC. This ablation study assesses the impact of cat type versus add type versus none type on the model's classification capability.

| | F1 | Precision | Recall | AUPRC | ROC-AUC |
|---|---|---|---|---|---|
| CCNETS (add) | **0.829** | **0.847** | **0.812** | **0.808** | **0.928** |
| CCNETS (cat) | 0.758 | 0.730 | 0.805 | 0.788 | 0.902 |

| CCNETS (none) | 0.371 | 0.377 | 0.790 | 0.348 | 0.656 |
|---|---|---|---|---|---|

Note: Boldface marks the best performance in each column.

This trend was even more pronounced in the AI4I 2020 Predictive Maintenance dataset, as detailed in Table 5. Here, the add type significantly outperformed all other variants across the entire set of metrics ($p < 0.05$).

**Table 5.** Performance comparison of CCNETS configurations with different Joint Types on the AI4I 2020 Predictive Maintenance dataset. The evaluation metrics include Precision, Recall, F1-score, ROC-AUC, and AUPRC. This ablation study assesses the impact of cat type versus add type versus none type on the model's classification capability.

| | F1 | Precision | Recall | AUPRC | ROC-AUC |
|---|---|---|---|---|---|
| CCNETS (add) | **0.984** | **0.997** | **0.973** | **0.977** | **0.987** |
| CCNETS (cat) | 0.914 | 0.894 | 0.942 | 0.954 | 0.976 |
| CCNETS (none) | 0.496 | 0.520 | 0.895 | 0.738 | 0.862 |

Note: Boldface marks the best performance in each column.

Consequently, we adopt the add type as the default setting for CCNETS. Beyond its empirical superiority, the add type offers distinct structural advantages over the second-best performing cat type. By utilizing element-wise addition rather than concatenation, it maintains a lower parameter count, which not only enhances computational efficiency but also mitigates the potential for overfitting.

### 3.5. Experiment 3: Impact of Data Augmentation Scale

The third experiment was conducted to provide a concrete interpretability analysis by visualizing the high-dimensional latent representations. In the CCNETS architecture, the Explainer and Reasoner collaboratively extract features to construct the latent space, after which the Reasoner performs the final classification. To validate whether this cooperative feature extraction process effectively learns discriminative representations despite extreme class imbalance, we employed t-SNE to project the latent vectors of correctly classified samples into a two-dimensional space.

Figure 11 illustrates the resulting feature manifolds for both the Credit Card Fraud Detection dataset (a) and the AI4I 2020 Predictive Maintenance dataset (b). In both visualizations, the majority class samples (Normal, light blue) form a dominant, continuous manifold, whereas the minority class samples (Fraud/Failure, orange) are mapped to distinct clusters.

Specifically, in Figure 11a, the observed overlaps reflect the inherent difficulty of fraud detection. Unlike the AI4I dataset (Figure 11b), where machine failures follow deterministic physical laws with lower feature dimensionality (14 features), credit card fraud presents a dual challenge: (1) the adversarial nature of stochastic human behaviors that mimic normal transactions, and (2) the higher topological complexity of the 30-dimensional feature space. This combination makes the underlying data manifold significantly more intricate, limiting the extent to which the classes can be linearly separated even when projected into the 256-dimensional latent space, resulting in unavoidable overlaps in the t-SNE visualization. In contrast, Figure 11b exhibits a sharper separation, where machine failure instances are clearly isolated from the normal operational data. This effective topological disentanglement demonstrates that the joint feature extraction by the Explainer and Reasoner has successfully captured high-level semantic features. Consequently, these visualizations serve as qualitative evidence of the model's interpretability, confirming that the Reasoner utilizes these robust representations to establish a discriminative decision boundary, effectively handling both complex overlapping boundaries and distinct anomaly patterns.

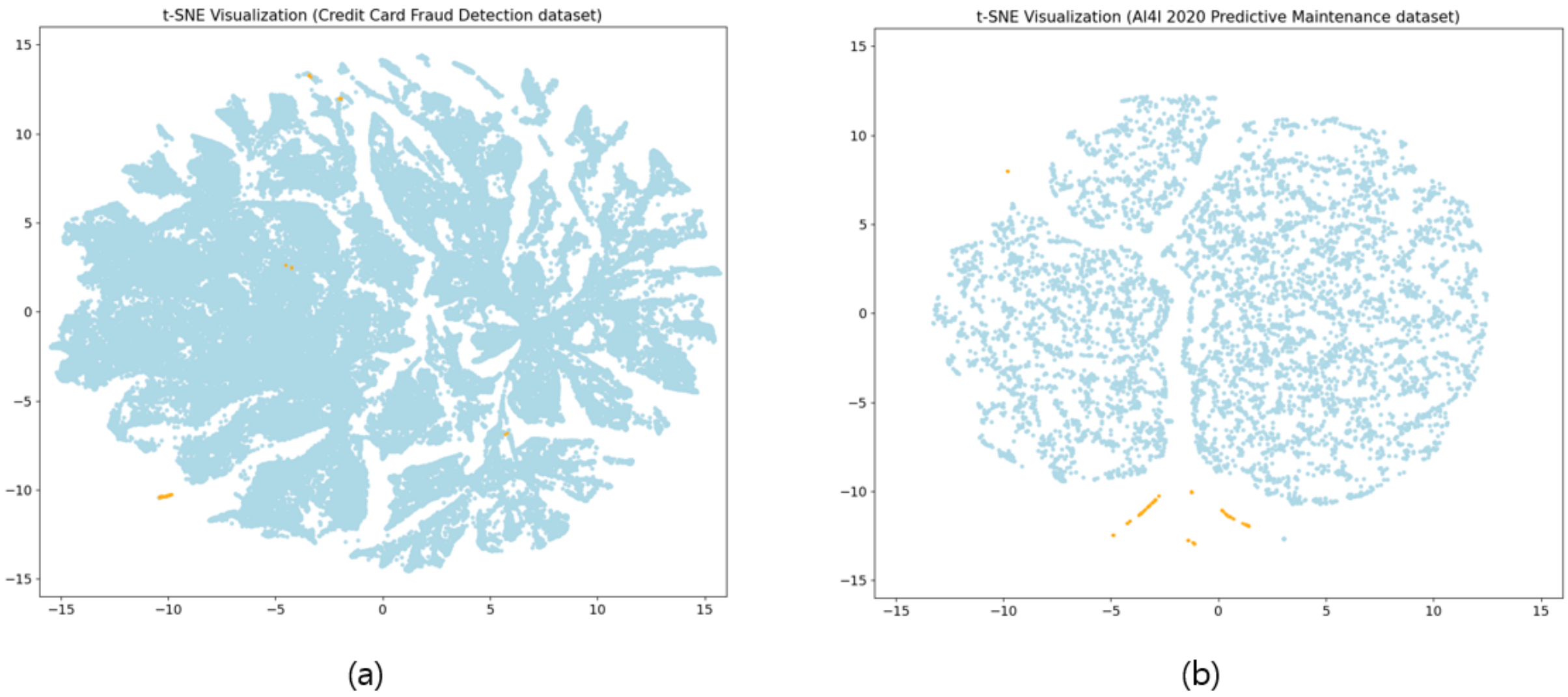

**Figure 11.** t-SNE visualization of latent representations jointly constructed by the Explainer and Reasoner for correctly classified samples. The plots illustrate the spatial distribution of the majority class (Normal, light blue) and the minority class (Fraud/Failure, orange) in the high-dimensional latent space. (a) Visualization for the Credit Card Fraud Detection dataset. While minor overlaps reflect the inherent complexity of subtle fraud patterns, the majority of fraudulent transactions form compact clusters distinct from the normal manifold. (b) Visualization for the AI4I 2020 Predictive Maintenance dataset, exhibiting a sharply separated distribution where failure instances form isolated groups. These results demonstrate that the cooperative feature extraction effectively establishes a discriminative decision boundary.

### 3.6. Experiment 4: Effectiveness of Generated Data

The fourth experiment was designed to evaluate whether data generated by CCNETS could enhance the performance of supervised learning models compared to models trained on the original dataset. In this experiment, we utilized a multilayer perceptron (MLP) as the supervised learning classifier and compared the outcomes across three different training data sources: original data, generation data, and reconstruction data produced by CCNETS.

Table 6 summarizes the performance comparison in terms of F1 score, precision, and recall. The model trained on generation data achieved the highest F1 score (0.8111), outperforming both the model trained on original data (0.7788) and the model trained on reconstruction data (0.7895). This indicates that the data generated by the CCNETS Producer significantly improved the model's ability to generalize.

**Table 6.** Supervised learning performance using original and CCNETS-generated data. This table compares the F1 score, precision, and recall when training an MLP classifier on original data, generated data, and reconstructed data. It highlights the enhancement achieved through data generated by the CCNETS Producer.

| | **f1** | **precision** | **Recall** |
|---|---|---|---|
| Original with MLP | 0.7788 | 0.8548 | 0.7153 |
| Generation with MLP | 0.8111 | 0.8690 | 0.7604 |
| Reconstruction with MLP | 0.7895 | 0.8607 | 0.7292 |

Precision and recall values also followed a similar pattern. The generation data model achieved a precision of 0.8690 and a recall of 0.7604, whereas the original data model achieved a precision of 0.8548 and a recall of 0.7153. These results suggest that synthetic data generated by CCNETS not only maintained high precision but also considerably improved the recall, which is critical in imbalanced classification tasks such as fraud detection.

In addition to the performance metrics, the learning dynamics were analyzed through Figure 11. Figure 11a presents the training loss curves for models trained on original, generation, and reconstruction data. It was observed that models trained with generation and reconstruction data exhibited faster and more stable convergence compared to those trained on the original dataset. Figure 11b illustrates the testing loss curves, confirming that models trained with generated data demonstrated superior generalization ability. Notably, the generation data model consistently achieved lower loss values during the testing phase, indicating enhanced robustness against overfitting.

These findings collectively demonstrate that CCNETS-generated data can effectively serve as high-quality training material, improving both the learning efficiency and predictive performance of supervised learning models, especially in domains characterized by significant class imbalance.

These results demonstrate that CCNETS not only enhances minority class detection but also embodies the soft computing principle of adaptive optimization under real-world data uncertainty.

### 3.7. Experiment 5: Impact of Data Augmentation Scale

The fifth experiment was conducted to investigate the impact of significantly increasing the volume of training data generated by CCNETS. Specifically, we compared the performance of supervised learning models trained on a single set of generated data versus models trained on data augmented tenfold through iterative generation processes.

Table 7 presents the performance comparison based on F1 score, precision, and recall. The model trained on the tenfold augmented dataset achieved a slightly higher F1 score (0.8133) compared to the model trained on the single generation dataset (0.8111). Precision and recall also showed slight improvements, with the tenfold model achieving a precision of 0.8696 and

a recall of 0.7639.

**Table 7.** Impact of tenfold data augmentation using CCNETS-generated data. This table presents the comparison results when the MLP is trained with data generated once versus data augmented to ten times the original size. A slight performance gain is observed with extensive data augmentation.

| | **f1** | **Precision** | **Recall** |
|---|---|---|---|
| Generation with MLP | 0.8111 | 0.8690 | 0.7604 |
| Generation X 10 | 0.8133 | 0.8696 | 0.7639 |

Although the improvements were relatively modest, the consistent enhancement across all metrics suggests that increasing the quantity of generated data can contribute to better model generalization and robustness. This is particularly important in real-world applications where collecting large amounts of labeled data is challenging.

These observations confirm that large-scale data augmentation can improve model performance.

In conclusion, while the performance gains from data augmentation were incremental, they validate the utility of CCNETS in addressing data scarcity issues through effective synthetic data generation, offering a scalable solution for enhancing supervised learning in data-constrained environments.

### 3.8. Summary of Key Findings

This study conducted a systematic evaluation of CCNETS through a series of experiments to validate its effectiveness in highly imbalanced classification tasks, including fraud detection and industrial fault detection. The results across all experiments consistently demonstrated the strengths of the proposed framework.

In Experiment 1, CCNETS was benchmarked against baselines including autoencoderbased detectors and DeepFM variants. Across both the Credit Card Fraud and AI4I datasets, CCNETS demonstrated superior discriminative capability, achieving statistically significant improvements in F1-score, AUPRC ($p < 0.05$). These results indicate that CCNETS not only excels at identifying rare instances by balancing precision and recall but also maintains robust generalization capabilities compared to traditional methods.

In Experiment 2, an ablation study on joint types confirmed the add configuration as the optimal strategy. It consistently demonstrated statistically significant superiority ($p < 0.05$) across key metrics in both the Credit Card Fraud Detection and AI4I datasets. This validates the selection of the add type as the default, as it offers distinct advantages over the second-best cat type: by utilizing element-wise addition, it ensures lower parameter complexity and higher computational efficiency, effectively mitigating the risk of overfitting while maintaining top-tier performance.

In Experiment 3, qualitative evidence of model interpretability was established through t-SNE

visualization of high-dimensional latent representations. The analysis revealed effective topological disentanglement, where minority class instances formed distinct, compact clusters separate from the majority manifold in both datasets. While the Credit Card Fraud dataset exhibited minor overlaps reflecting inherent task complexity, the AI4I dataset demonstrated sharp separation. These visualizations confirm that the collaborative feature extraction by the Explainer and Reasoner successfully captures high-level semantic features, enabling the construction of robust discriminative boundaries even in highly imbalanced scenarios.

In Experiment 4, we evaluated the impact of training supervised learning models with data generated by CCNETS. Models trained on generation data outperformed those trained on original or reconstruction data, achieving higher F1 scores, precision, and recall. This demonstrated that CCNETS can generate high-quality synthetic data that enhances model training, leading to superior generalization.

In Experiment 5, we assessed the effect of scaling up the generated dataset by tenfold. Although the performance improvements were incremental, consistent gains in F1 score, precision, and recall were observed. Additionally, faster convergence and lower testing losses were noted, confirming that larger quantities of generated data can further stabilize and enhance model training.

Throughout the experiments, Figures 9, 10 and 12 illustrated stable loss reductions during training and testing phases, further validating the learning effectiveness and generalization ability of CCNETS.

Overall, the experimental results confirm that CCNETS effectively addresses the limitations of traditional generative models while offering a robust solution to data scarcity through high-quality synthetic data generation. These outcomes establish CCNETS as a scalable and effective framework for improving supervised learning performance in scenarios marked by severe class imbalance. By combining generation and inference within a unified, modular architecture under uncertainty, CCNETS embodies the core principles of hybrid soft computing.

Although this study primarily compared CCNETS against an autoencoder-based baseline and DeepFM variants, future work will extend the evaluation to include widely used imbalance handling techniques such as SMOTE, ADASYN, and cost-sensitive learning. Nevertheless, the current results suggest that CCNETS's feedback-driven generation mechanism confers significant advantages over conventional static oversampling strategies, particularly in terms of adaptability and targeted minority class reinforcement.

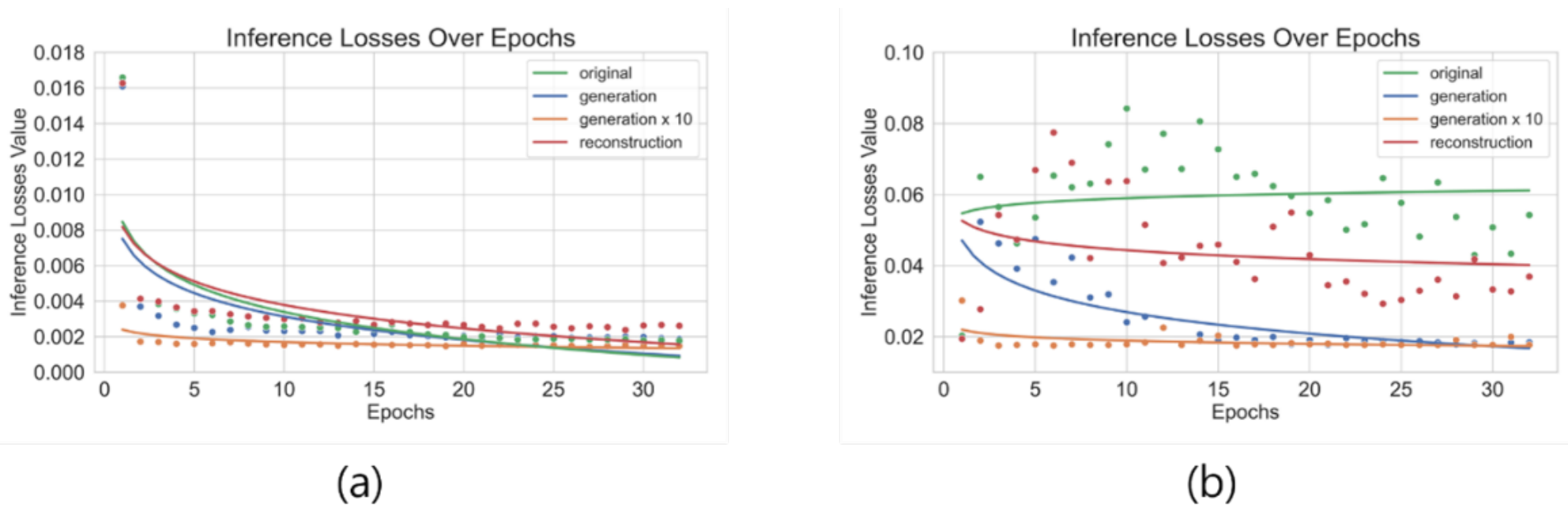

**Figure 12.** Loss trends of MLP models trained with original and CCNETS-generated data. This figure compares inference loss curves during MLP training and testing using four types of data: original,

generation, generation ×10, and reconstruction data. Faster convergence of models trained with CCNETS-generated data suggests improved learning efficiency. (a) Inference loss during the training process, exhibiting an overall decreasing trend. The R 2 scores are as follows: original data at 0.5309, generation data at 0.4292, generation × 10 data at 0.4251, and reconstruction data at 0.4615. (b) Inference loss during the testing process, also indicating an overall decreasing trend. The R 2 scores for this phase are original data at 0.016, generation data at 0.5379, generation × 10 data at 0.2331, and reconstruction data at 0.052.

## 4. Discussion

### 4.1. Overall Evaluation of CCNETS

This study presented Causal Cooperative Networks (CCNETS), a modular and causally linked learning framework that addresses the core challenges of imbalanced classification. The need for such a framework stem from longstanding limitations in traditional generative models such as GANs[5], DCGAN[6], and StyleGAN[8]. While these models excel at generating realistic data, they fall short when directly applied to classification tasks in severely imbalanced contexts. Prior work has shown that generative augmentation techniques often suffer from instability and weak synergy with discriminative models[12,21,23,37], limiting their overall impact in classification pipelines.

CCNETS responds to these limitations by decoupling the critical learning tasks—feature abstraction, label inference, and data generation—into three dedicated modules: the Explainer, Reasoner, and Producer. This structural decomposition aligns with the modular learning principles advocated in prior research, where distributed representation learning improved model stability and interpretability[11,13,16,38]. Unlike single-network approaches that conflate generation and classification, CCNETS explicitly separates responsibilities while enabling communication through a causal learning loop, resulting in more stable optimization and clearer function attribution.

Experimental validation on two distinct, highly imbalanced domains—credit card fraud detection and industrial fault detection—demonstrated CCNETS's broad practical utility. The Reasoner module consistently achieved superior F1-scores and AUPRC compared to conventional autoencoder-based methods and DeepFM variants, driven by improved recall of the minority class. This aligns with the priorities of anomaly detection literature, where recall is considered a vital metric in safety-critical environments[6,12,39]. CCNETS thus meets a key objective of real-world anomaly detection systems: reducing false negatives. This capability is critical not only in financial systems, where missed fraud leads to direct monetary losses and reputational damage, but also in predictive maintenance scenarios, where failing to detect faults can result in catastrophic equipment failure and operational downtime. Consequently, the versatility demonstrated across these distinct domains suggests that CCNETS holds significant potential for generalization to a wider range of high-stakes, class-imbalanced applications.

Beyond classification accuracy, the Producer module contributed significantly by generating high-quality, semantically consistent data that accelerated convergence and improved generalization. Similar benefits of generative augmentation have been documented across domains such as speech recognition[16], medical imaging[10], and autonomous driving[11], where GAN-based frameworks have enabled synthetic expansion of limited datasets. In particular, the Producer in CCNETS achieved performance comparable to specialized GAN variants that are typically fine-tuned for specific domains[7,17,24].

One of the most notable outcomes of this study is CCNETS's ability to generalize effectively from limited real-world data. While conventional augmentation strategies often rely on abundant training samples to learn useful latent representations[25,26], CCNETS instead benefits from its causal coupling of data generation and label inference. This feedback loop enables the system to generate targeted synthetic examples that directly support the Reasoner's weaknesses, leading to improved generalization without the need for large-scale datasets. This characteristic makes CCNETS particularly relevant for domains where data acquisition is constrained—such as finance, cybersecurity, and clinical medicine—where collecting representative, labeled, minority-class data is costly and complex.

Although CCNETS draws light conceptual inspiration from biologically modular systems[40], it does not aim to replicate any cognitive process. Instead, its structure—consisting of interpretable and functional abstraction, inference, and generation modules—echoes the architectures explored in brain decoding studies[29,30]. These analogies suggest potential for interdisciplinary interpretation but serve mainly to support the modular, soft-computing-friendly architecture proposed in this work.

In summary, CCNETS offers a substantial advancement over previous generativediscriminative models in the context of imbalanced learning. By leveraging causal cooperation between modules, adaptive data generation, and explicit modular decomposition, it introduces a robust framework that directly aligns with the principles of hybrid soft computing—modularity, flexibility, adaptability, and robustness under uncertain conditions.

### 4.2. Analysis of Strengths and Limitations

The CCNETS framework delivers multiple strengths that distinguish it from conventional solutions for imbalanced data classification.

First, its explicit modular design—with separated roles for the Explainer, Reasoner, and Producer—enables task specialization and architectural clarity. This separation overcomes the coupling issues common in prior unified models, where mixing generative and discriminative objectives often led to training instability[12,13,21]. Each component in CCNETS is independently optimizable, improving training dynamics while maintaining functional integration. The Explainer captures high-dimensional, latent features in line with hierarchical abstraction practices in deep networks [1,6]; the Reasoner applies these features jointly with input data to make label predictions, consistent with enhanced discriminative models that fuse feature-label learning[11,19]; and the Producer contributes through synthetic generation, reinforcing model robustness—a benefit repeatedly demonstrated in the data-augmentation literature[9,10,16,17,25].

Second, CCNETS provides interpretability and transparency, qualities often lacking in typical "black-box" deep learning models. Unlike conventional single-network architectures where the internal decision logic remains opaque, CCNETS allows for qualitative validation of learned representations through latent space visualization. Our experiments demonstrated that, even in scenarios characterized by complex class overlaps, minority class instances were not dispersed as noise but mapped into distinct, compact clusters. This topological disentanglement confirms that the Explainer and Reasoner are not merely memorizing data but are effectively capturing high-level semantic features to establish robust decision boundaries.

Third, generalization under data scarcity is a defining advantage of CCNETS. Unlike traditional oversampling techniques or GAN-based augmentation methods[5,24], CCNETS's closed-loop feedback mechanism allows it to adaptively generate minority class data in response to classification gaps. The resulting synergy enhances learning efficiency and reduces overfitting—a pattern supported by our experiments and echoed in prior domain-specific studies on medical imaging[10] and environmental sound classification[28,41].

Fourth, scalability is demonstrated in our augmentation experiments. Increasing the amount of Producer-generated synthetic data tenfold resulted in consistent performance gains, indicating that CCNETS can scale to accommodate different data budgets. This trend aligns with previous research on data amplification as a means of reinforcing model resilience in skewed class distributions[25–27].

Nonetheless, several limitations must be acknowledged. One key constraint is the framework's sensitivity to underlying neural architectures. As our ablation studies showed, the choice of DeepFM[42] for the Reasoner and ResMLP[43] for the Producer influenced final performance. Inappropriate selection or mismatched module complexity may lead to suboptimal convergence or task imbalance, suggesting a need for careful architecture tuning per application.

Another issue concerns the training stability and quality of the synthetic data produced by the Producer. As noted in earlier work [7,14], generative models are prone to instability and may produce noisy or unrealistic samples when poorly optimized. While CCNETS's causal integration partially mitigates this, it does not entirely eliminate the dependence on robust generator training. Simply increasing data volume is not a guarantee of quality unless the Producer is adequately tuned and monitored.

Finally, the modular design of CCNETS introduces added complexity in hyperparameter coordination. Each module involves its own set of tunable parameters, and effective synchronization is essential to maintain stable cooperation. Improper alignment of learning rates or update schedules may disrupt the collaborative dynamics and degrade system-level performance.

Nevertheless, CCNETS offers a robust and flexible solution for imbalanced classification. It is clearly defined module architecture, integrated generative-discriminative strategy, and scalability make it applicable across a wide range of domains. Achieving optimal performance, however, requires careful architectural configuration, vigilant supervision of the generative process, and coherent optimization across modules.

In contrast to traditional oversampling techniques such as SMOTE or ADASYN, which create synthetic data independently of classifier behavior, CCNETS enables a feedback driven generation process that adaptively focuses on the classifier's deficiencies. This interaction loop offers a structural advantage by producing synthetic examples that are causally aligned with learning objectives, rather than being randomly interpolated from feature space.

### 4.3. Analysis of Strengths and Limitations

The findings of this study have far-reaching implications for intelligent systems operating under data scarcity and imbalance. CCNETS introduces a new avenue for integrating synthetic data generation and causal label inference in a soft computing context—unlike traditional GAN-based augmentation frameworks that treat generation and classification as disconnected stages[5,9,12,16,24]. By creating a feedback-aligned architecture, CCNETS addresses a fundamental gap in ensuring that generated data serve the direct needs of the classifier[11,21].

The most important practical implication is that structured, causally informed data generation, when tightly integrated with label prediction, can significantly improve classifier robustness and generalization. Our empirical results reaffirm this, demonstrating consistent improvements across metrics, especially recall—highlighting the framework's utility in domains where minority-class detection is critical. This observation aligns with other reports of data augmentation improving model stability in data-scarce environments[10,17,25,44].

CCNETS's scalability further enhances its relevance in operational contexts. As seen in applications such as hyperspectral imaging[17,26] and sound classification[28,41], modest increases in high-quality synthetic data yield substantial gains. CCNETS supports such extensions natively and offers a cost-effective alternative to collecting large-scale labeled datasets, especially in regulated or privacy-constrained domains.

Several research opportunities emerge from this study. First, the architecture could benefit from further exploration of internal network designs. Alternatives to MLPs and DeepFM—such as transformers or graph neural networks—may improve reasoning and feature abstraction capacities[15,19], especially on structured or relational data.

Second, enhanced generation strategies could improve the Producer's output quality. Techniques such as progressive GANs[8] and self-supervised representation learning[14] may offer richer latent spaces and more stable training, especially in low-data regimes.

Third, CCNETS's design—while inspired conceptually by modular cognition[39]—could be extended through interdisciplinary integration. Future work might incorporate biologically inspired dynamics such as adaptive attention or online feedback modulation based on cognitive neuroscience principles[29–31].

Finally, domain-wide validation is crucial for future adoption. Applying CCNETS across real-time fraud detection systems, medical diagnostics, and industrial fault detection will provide further insight into its generalizability and practical robustness. Similar cross domain extensions have proven successful in recent GAN-based applications[11,13,23,27].

In summary, CCNETS offers more than just a method for imbalanced classification—it lays the groundwork for a new class of modular, adaptive, and interpretable intelligent systems rooted in the principles of soft computing. Future work at the intersection of generative modeling, causal inference, and bio-inspired learning could extend its impact well beyond its initial domain.

## 5. Conclusion

This study introduced Causal Cooperative Networks (CCNETS), a modular and adaptive learning framework meticulously designed to overcome the persistent challenges of classification within severely imbalanced and data-scarce environments. Building upon foundational advances in generative modeling[5,8,9] and robust strategies for pattern recognition[16,17,25], CCNETS establishes a new paradigm by integrating generative and discriminative learning within a structured, functionally decomposed architecture. In contrast to static data augmentation techniques such as SMOTE or conventional GAN-based preprocessing pipelines—which often operate independently of classifier feedback and suffer from instability[11,21]—CCNETS synchronizes generation and inference through a synchronous, closed-loop causal feedback system. This architecture allows the classifier to act as a functional cause that dynamically guides the generator, ensuring that synthetic data directly addresses the model's specific predictive weaknesses.

Rather than employing a monolithic structure, CCNETS assigns specialized, cooperative roles to its three core modules: the Explainer for extracting meaningful latent features; the Reasoner for inferring labels using both raw and latent inputs; and the Producer for synthesizing data to reinforce minority class representation. Extensive validation on the Credit Card Fraud Detection dataset (0.17% fraud cases) and the AI4I 2020 Predictive Maintenance dataset (3.39% failure cases) confirmed the framework's superior effectiveness. CCNETS consistently outperformed baseline models, including autoencoders and DeepFM variants, by achieving significantly higher F1-scores and AUPRC, particularly under limited data conditions. These results reinforce the critical value of causally aligned data augmentation in improving generalization performance when labeled data are scarce[10,17,26].

A defining strength of CCNETS is its scalability and operational flexibility. By modulating the volume and structure of generated data, the framework can be adaptively tailored to various deployment constraints without heavy reliance on additional real-world data collection. This adaptability makes it uniquely suited for high-stakes domains such as fraud detection, rare disease diagnosis, and industrial anomaly detection, where minority events are rare yet carry significant risk. Furthermore, while CCNETS draws conceptual inspiration from the functional abstraction and modularity of biological information-processing systems—echoing principles used in neuroscience-inspired image reconstruction and brain decoding[29,30]—it remains firmly rooted in engineering pragmatism for robust, real-world application.

In summary, CCNETS establishes a robust, scalable, and modular soft computing framework for tackling data imbalance and scarcity. Its architectural clarity and causal integration lay the groundwork for next-generation intelligent systems capable of autonomous reasoning under uncertainty. While this work focuses on architectural synergy, the inherent modularity of the design supports potential interpretability and granular diagnostic analysis, promising a fertile path for future extensions into explainable AI and adaptive attention mechanisms, ultimately facilitating the deployment of more reliable anomaly detection systems in critical industrial infrastructures.

**Author Contributions:** Conceptualization, H.P. and Y.C.; methodology, H.P. and Y.C.;

software, H.P.; validation, H.P. and Y.C.; formal analysis, H.P.; investigation, H.P. and Y.C.; data curation, H.P. and Y.C.; writing—original draft preparation, H.P. and Y.C.; writing—review and editing, H.P. and Y.C.; visualization, H.P. and Y.C.; supervision, H.K.; project administration, H.K. All authors have read and agreed to the published version of the manuscript.

**Funding:** This research was supported by the National Research Foundation of Korea government (RS-2023-00242528). This research was supported by the MSIT (Ministry of Science, ICT), Korea, under the National Program for Excellence in SW, supervised by the IITP (Institute of Information & communications Technology Planing & Evaluation) in 2026(2024-0-00018).

**Institutional Review Board Statement:** Not applicable.

**Informed Consent Statement:** Not applicable.

**Data Availability Statement:** The source code used for data processing and analysis is publicly available at https://github.com/bxailab/CCNETS (accessed on 25 January 2026).

**Conflicts of Interest:** The authors declare no conflicts of interest. The funders had no role in the design of the study; in the collection, analyses, or interpretation of data; in the writing of the manuscript; or in the decision to publish the results.

## References

[1] C.G. Turhan, H.S. Bilge, Recent Trends in Deep Generative Models: a Review, in: 2018 3rd International Conference on Computer Science and Engineering (UBMK), 2018: pp. 574–579. https://doi.org/10.1109/UBMK.2018.8566353.

[2] Z. Xing, O. Mehmood, W.A.P. Smith, Unsupervised anomaly detection with a temporal continuation, confidence-aware VAE-GAN, Pattern Recognit. 166 (2025) 111699. https://doi.org/https://doi.org/10.1016/j.patcog.2025.111699.

[3] N. Qiang, J. Gao, Q. Dong, H. Yue, H. Liang, L. Liu, J. Yu, J. Hu, S. Zhang, B. Ge, Y. Sun, Z. Liu, T. Liu, J. Li, H. Song, S. Zhao, Functional brain network identification and fMRI augmentation using a VAE-GAN framework, Comput. Biol. Med. 165 (2023) 107395. https://doi.org/https://doi.org/10.1016/j.compbiomed.2023.107395.

[4] Y. Han, Y. Liu, Q. Chen, Data augmentation in material images using the improved HP-VAE-GAN, Comput. Mater. Sci. 226 (2023) 112250. https://doi.org/https://doi.org/10.1016/j.commatsci.2023.112250.

[5] I.J. Goodfellow, J. Pouget-Abadie, M. Mirza, B. Xu, D. Warde-Farley, S. Ozair, A. Courville, Y. Bengio, Generative Adversarial Nets, in: Advances in Neural Information Processing Systems 27 (NIPS 2014), 2014: pp. 1–9. http://www.github.com/goodfeli/adversarial.

[6] A. Radford, L. Metz, S. Chintala, Unsupervised Representation Learning with Deep Convolutional Generative Adversarial Networks, ArXiv (2016). http://arxiv.org/abs/1511.06434.

[7] T.R. Shaham, T. Dekel, T. Michaeli, SinGAN: Learning a Generative Model From a Single Natural Image, in: 2019 IEEE/CVF International Conference on Computer Vision (ICCV), 2019: pp. 4569–4579. https://doi.org/10.1109/ICCV.2019.00467.

[8] T. Karras, S. Laine, T. Aila, A Style-Based Generator Architecture for Generative Adversarial Networks, ArXiv (2019). http://arxiv.org/abs/1812.04948.

[9] N. Sasipriyaa, P. Natesan, E. Gothai, G. Madhesan, E. Madhumitha, K. V Mithun, Recognition of Tamil handwritten characters using Scrabble GAN, in: 2023 International Conference on Computer Communication and Informatics (ICCCI), 2023: pp. 1–5. https://doi.org/10.1109/ICCCI56745.2023.10128564.

[10] Yasamin Kowsari, Seyed Javad Mahdavi Chabok, Mohammad Hossein Moattar, Classification of Pulmonary Images By Using Generative Adversarial Networks, in: 8th Iranian Joint Congress on Fuzzy and Intelligent Systems (CFIS) : September 2-4, 2020 - Ferdowsi University of Mashhad, 2020: pp. 133–137.

[11] R. Sampath Kumar, V.P. Krishnamurthy, V. Podile, G. Yamini Priyanka, V. Neha, Generative Adversarial Networks to Improve the Nature of Training in Autonomous Vehicles, in: 2023 International Conference on Disruptive Technologies, ICDT 2023, Institute of Electrical and Electronics Engineers Inc., 2023: pp. 161–164. https://doi.org/10.1109/ICDT57929.2023.10151288.

[12] Shuai Zheng, Chetan Gupta, Discriminant Generative Adversarial Networks with Its Application to Equipment Health Classification, in: ICASSP 2020 - 2020 IEEE International Conference on Acoustics, Speech and Signal Processing (ICASSP), IEEE, 2020: pp. 3067–3071.

[13] Y. Shi, Q. Li, X.X. Zhu, Building Footprint Generation Using Improved Generative Adversarial Networks, IEEE Geoscience and Remote Sensing Letters 16 (2019) 603–607. https://doi.org/10.1109/LGRS.2018.2878486.

[14] X. Chen, R. Roberts, Z. Liu, W. Tong, A generative adversarial network model alternative to animal studies for clinical pathology assessment, Nat. Commun. 14 (2023). https://doi.org/10.1038/s41467-023-42933-9.

[15] Q. Zhang, J. Yang, X. Zhang, T. Cao, Generating Adversarial Examples in Audio Classification with Generative Adversarial Network, in: 2022 7th International Conference on Image, Vision and Computing, ICIVC 2022, Institute of Electrical and Electronics Engineers Inc., 2022: pp. 848–853. https://doi.org/10.1109/ICIVC55077.2022.9886154.

[16] P. Shastri, C. Patil, P. Wanere, S. Mahajan, A. Bhatt, Adversarial Synthesis based Data Augmentation for Speech Classification, in: 2022 International Conference on Signal and Information Processing, IConSIP 2022, Institute of Electrical and Electronics Engineers Inc., 2022. https://doi.org/10.1109/ICoNSIP49665.2022.10007491.

[17] Y. Zhan, Y. Wang, X. Yu, Semisupervised hyperspectral image classification based on generative adversarial networks and spectral angle distance, Sci. Rep. 13 (2023). https://doi.org/10.1038/s41598-023-49239-2.

[18] Z. Yu, W. Cui, Robust hyperspectral image classification using generative adversarial networks, Inf. Sci. (N. Y). 666 (2024) 120452. https://doi.org/https://doi.org/10.1016/j.ins.2024.120452.

[19] H. Gao, H. Zhang, X. Yang, W. Li, F. Gao, Q. Wen, Generating natural adversarial examples with universal perturbations for text classification, Neurocomputing 471 (2022) 175–182. https://doi.org/10.1016/j.neucom.2021.10.089.

[20] A.A.N.D.V.-C.C.A.N.D.M.-A.C. Gutiérrez Benítez Rodrigo AND Segura Navarrete, Guide for the application of the data augmentation approach on sets of texts in Spanish for sentiment and

emotion analysis, PLoS One 19 (2024) 1–24. https://doi.org/10.1371/journal.pone.0310707.

[21] H.-S. Choi, D. Jung, S. Kim, S. Yoon, Imbalanced Data Classification via Cooperative Interaction Between Classifier and Generator, IEEE Trans. Neural Netw. Learn. Syst. 33 (2022) 3343–3356. https://doi.org/10.1109/TNNLS.2021.3052243.

[22] S.D. Wickramaratne, M.D.S. Mahmud, LSTM based GAN Networks for Enhancing Ternary Task Classification Using fNIRS Data, in: 2021 43rd Annual International Conference of the IEEE Engineering in Medicine & Biology Society (EMBC), 2021: pp. 1043–1046. https://doi.org/10.1109/EMBC46164.2021.9630000.

[23] A.N. Dar, R. Rastogi, MLGAN: Addressing Imbalance in Multilabel Learning Using Generative Adversarial Networks, in: 2023 International Conference on Emerging Techniques in Computational Intelligence (ICETCI), 2023: pp. 324–331. https://doi.org/10.1109/ICETCI58599.2023.10331105.

[24] Y.-W. Lu, K.-L. Liu, C.-Y. Hsu, Conditional Generative Adversarial Network for Defect Classification with Class Imbalance, in: 2019 IEEE International Conference on Smart Manufacturing, Industrial & Logistics Engineering (SMILE), 2019: pp. 146–149. https://doi.org/10.1109/SMILE45626.2019.8965320.

[25] C. Liu, X. Wang, K. Wu, J. Tan, F. Li, W. Liu, Oversampling for Imbalanced Time Series Classification Based on Generative Adversarial Networks, in: 2018 IEEE 4th International Conference on Computer and Communications (ICCC), 2018: pp. 1104–1108. https://doi.org/10.1109/CompComm.2018.8780808.

[26] V.-E. Neagoe, P. Diaconescu, CNN Hyperspectral Image Classification Using Training Sample Augmentation with Generative Adversarial Networks, in: 2020 13th International Conference on Communications (COMM), 2020: pp. 515–519. https://doi.org/10.1109/COMM48946.2020.9142021.

[27] X. Zhu, Y. Liu, J. Li, T. Wan, Z. Qin, Emotion classification with data augmentation using generative adversarial networks, in: Lecture Notes in Computer Science (Including Subseries Lecture Notes in Artificial Intelligence and Lecture Notes in Bioinformatics), Springer Verlag, 2018: pp. 349–360. https://doi.org/10.1007/978-3-319-93040-4_28.

[28] Aswathy Madhu, Suresh Kumaraswamy, Data Augmentation Using Generative Adversarial Network for Environmental Sound Classification, in: 27th EUSIPCO 2019 : European Signal Processing Conference : A Coruña, Spain, September 2-6, 2019, 2019.

[29] K. Seeliger, U. Güçlü, L. Ambrogioni, Y. Güçlütürk, M.A.J. van Gerven, Generative adversarial

networks for reconstructing natural images from brain activity, Neuroimage 181 (2018) 775–785. https://doi.org/10.1016/j.neuroimage.2018.07.043.

[30] G. Gaziv, R. Beliy, N. Granot, A. Hoogi, F. Strappini, T. Golan, M. Irani, Self-supervised Natural Image Reconstruction and Large-scale Semantic Classification from Brain Activity, Neuroimage 254 (2022). https://doi.org/10.1016/j.neuroimage.2022.119121.

[31] Y. Miyawaki, H. Uchida, O. Yamashita, M. aki Sato, Y. Morito, H.C. Tanabe, N. Sadato, Y. Kamitani, Visual Image Reconstruction from Human Brain Activity using a Combination of Multiscale Local Image Decoders, Neuron 60 (2008) 915–929. https://doi.org/10.1016/j.neuron.2008.11.004.

[32] T. Dado, Y. Güçlütürk, L. Ambrogioni, G. Ras, S. Bosch, M. van Gerven, U. Güçlü, Hyperrealistic neural decoding for reconstructing faces from fMRI activations via the GAN latent space, Sci. Rep. 12 (2022). https://doi.org/10.1038/s41598-021-03938-w.

[33] T. Naselaris, R.J. Prenger, K.N. Kay, M. Oliver, J.L. Gallant, Bayesian Reconstruction of Natural Images from Human Brain Activity, Neuron 63 (2009) 902–915. https://doi.org/10.1016/j.neuron.2009.09.006.

[34] S. Gopalakrishnan, P.R. Singh, Y. Yazici, C.S. Foo, V. Chandrasekhar, A.M. Ambikapathi, Classify and generate: Using classification latent space representations for image generations, Neurocomputing 471 (2022) 296–334. https://doi.org/10.1016/j.neucom.2021.10.090.

[35] J. Park. Framework for causal learning of neural networks, WIPO Patent, WO2022164299A1, 04 August 2022. https://patents.google.com/patent/WO2022164299A1/.

[36] S. Schoenmakers, M. Barth, T. Heskes, M. van Gerven, Linear reconstruction of perceived images from human brain activity, Neuroimage 83 (2013) 951–961. https://doi.org/10.1016/j.neuroimage.2013.07.043.

[37] C. Li, X. Wang, B. Miao, C. Xie, Z. Wang, Y. Zhu, An Efficient Framework for Enhancing Discriminative Models via Diffusion Techniques, Proceedings of the AAAI Conference on Artificial Intelligence 39 (2025) 4670–4678. https://doi.org/10.1609/aaai.v39i5.32493.

[38] V. Artiushenko, R. Reider, T. Reggelin, S. Lang, Advancing robotic systems with Ensemble and Modular Deep Learning: research idea and framework, Procedia Comput. Sci. 253 (2025) 2557–2566. https://doi.org/https://doi.org/10.1016/j.procs.2025.01.315.

[39] S. Chen, H. Min, Y. Fang, X. Wu, B. Li, X. Zhao, Uncertainty-aware Sensor Data Anomaly Detection for Autonomous Vehicles, in: 2024 IEEE Intelligent Vehicles Symposium (IV), 2024: pp. 478–483. https://doi.org/10.1109/IV55156.2024.10588587.

[40] A. Ororbia, D. Kifer, The neural coding framework for learning generative models, Nat. Commun. 13 (2022) 2064. https://doi.org/10.1038/s41467-022-29632-7.

[41] B. Bahmei, E. Birmingham, S. Arzanpour, CNN-RNN and Data Augmentation Using Deep Convolutional Generative Adversarial Network for Environmental Sound Classification, IEEE Signal Process. Lett. 29 (2022) 682–686. https://doi.org/10.1109/LSP.2022.3150258.

[42] H. Guo, R. Tang, Y. Ye, Z. Li, X. He, DeepFM: A Factorization-Machine based Neural Network for CTR Prediction, 2017.

[43] H. Touvron, P. Bojanowski, M. Caron, M. Cord, A. El-Nouby, E. Grave, G. Izacard, A. Joulin, G. Synnaeve, J. Verbeek, H. Jegou, ResMLP: Feedforward Networks for Image Classification with Data-Efficient Training, IEEE Trans. Pattern Anal. Mach. Intell. 45 (2023) 5314–5321. https://doi.org/10.1109/TPAMI.2022.3206148.

[44] Y. Wang, L. Deng, Enhanced Data Augmentation for Infrared Images With Generative Adversarial Networks Aided by Pretrained Models, IEEE Access 12 (2024) 176739–176750. https://doi.org/10.1109/ACCESS.2024.3491167.